\documentclass{ecai} 

\usepackage{latexsym}
\usepackage{amssymb}
\usepackage{amsmath}
\usepackage{amsthm}
\usepackage{booktabs}
\usepackage{enumitem}
\usepackage{graphicx}
\usepackage{color}
\usepackage{subcaption}
\usepackage{tabularx}

\newcommand{\BibTeX}{B\kern-.05em{\sc i\kern-.025em b}\kern-.08em\TeX}

\begin{document}


\begin{frontmatter}




\title{Revisiting Safe Exploration in Safe Reinforcement learning}


\author[A]{\fnms{David}~\snm{Eckel}\thanks{Corresponding Author. Email: eckeld@tf.uni-freiburg.de}}
\author[A]{\fnms{Baohe}~\snm{Zhang}\thanks{Corresponding Author. Email: zhangb@cs.uni-freiburg.de}}
\author[A]{\fnms{Joschka}~\snm{Bödecker}} 

\address[A]{University of Freiburg}


\begin{abstract}
Safe reinforcement learning (SafeRL) extends standard reinforcement learning with the idea of safety, where safety is typically defined through the constraint of the expected cost return of a trajectory being below a set limit.
However, this metric fails to distinguish how costs accrue, treating infrequent severe cost events as equal to frequent mild ones, which can lead to riskier behaviors and result in unsafe exploration. We introduce a new metric, expected maximum consecutive cost steps (EMCC), which addresses safety during training by assessing the severity of unsafe steps based on their consecutive occurrence. This metric is particularly effective for distinguishing between prolonged and occasional safety violations. We apply EMMC in both on- and off-policy algorithm for benchmarking their safe exploration capability. Finally, we validate our metric through a set of benchmarks and propose a new lightweight benchmark task, which allows fast evaluation for algorithm design.
\end{abstract}
\end{frontmatter}


\section{Introduction}
Defining a reward function for Reinforcement Learning is complex and requires significant expertise, particularly for real-world applications. Creating a single reward function that encapsulates all goals can be difficult and may result in sub-optimal policies due to varying importance of different reward components. Instead, formulating these tasks as constrained optimization problems can be more effective. For instance, in a heating system control scenario, it is more straightforward to formulate the thermal comfort as constraints and minimize the energy usage as reward rather than combining these objectives into one reward function.
To address these constrained optimization problems, SafeRL has been developed via formulating the problem as a constrained Markov decision problem (CMDP)~\citep{Altman1999CMDP} to ensure that the control system adheres to critical safety constraints during both training and real-world deployment~\citep{Yinlam2018Lyapunov, Amodei2016AISafety}.

The trade-off between exploration and exploitation lies at the core of RL and plays the vital role for improving the data efficiency and overall performance. In the context of SafeRL, safety is crucial to prevent severe violations of constraints and potential harm to the agents and the environments. Thus, maintaining safety during the training and deployment of agents becomes a critical third dimension, alongside exploration and exploitation. However, this focus on safety can conflict with the need for exploration, particularly since SafeRL agents often interact with the environments without prior knowledge and must explore to learn safe behaviors. This scenario presents a significant dilemma: how can SafeRL algorithms balance safety with the necessity of exploration?

Many SafeRL benchmark work~\citep{ji2023safetygymnasium, liu2023datasets, zhao2023guard} have been carried out to compare the performance of different algorithms by looking into two metrics: expected cumulative return and costs of the final policy after training. 
\citet{Achiam2019BenchmarkingSE} proposes to use another metric for comparing safe exploration during training in form of the average cost over the entirety of training.
This metric directly corresponds to safety outcomes as a lower cost rate relates to less unsafe steps during training. However, all these metrics often do not adequately reflect the nuances of safe exploration. For instance, they may not differentiate between the severity of unsafe actions taken during exploration, thus potentially overlooking critical safety nuances.

\begin{figure}[h]
    \centering
    \begin{subfigure}[b]{0.45\columnwidth}
        \centering
        \includegraphics[width=\textwidth]{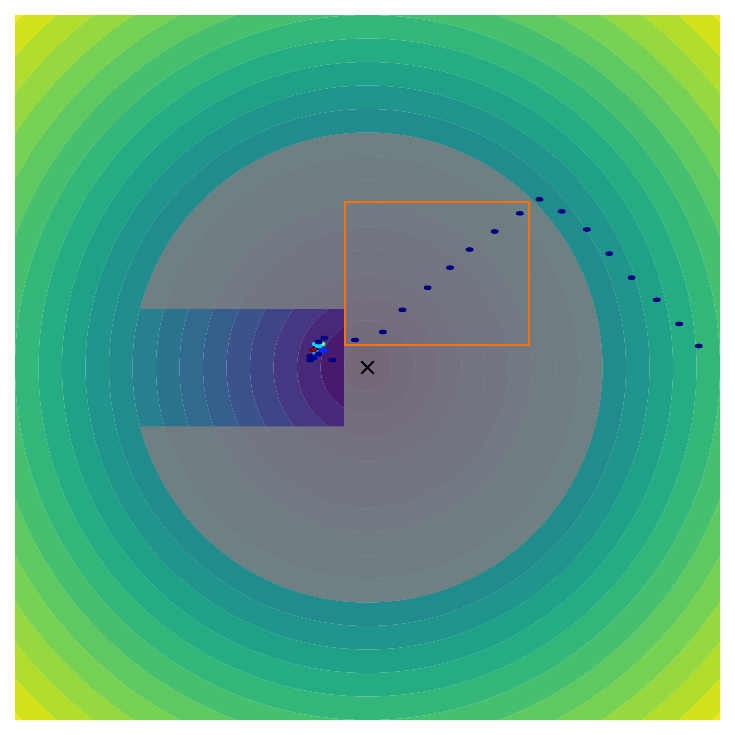} 
    \end{subfigure}
    \hfill
    \begin{subfigure}[b]{0.45\columnwidth}
        \centering
    \includegraphics[width=\textwidth]{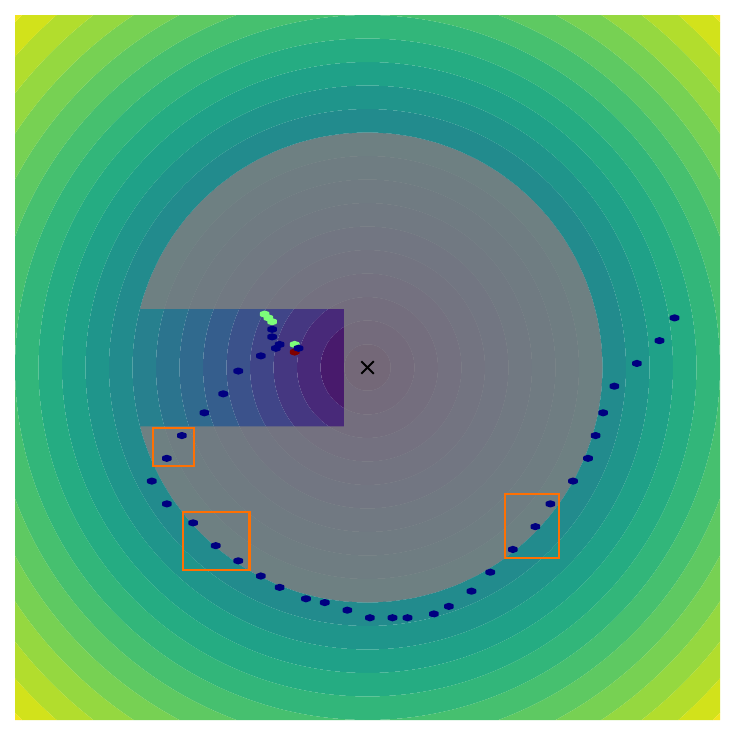}
    \end{subfigure}
    \caption{Starting from the right side, agents are asked to go as near as possible to the center of the circle while avoid stepping into the grey zone, which will give a step cost of $1$. Two trajectories (blue dots) with different numbers of consecutive cost steps. The \textbf{left} trajectory has a larger number of consecutive cost steps compared to the \textbf{right}. But both result in similar cumulative costs. The chains of consecutive unsafe steps are marked in the orange box.}
    \label{fig:EMCC_DifferentChains}
    \vspace{10pt}
\end{figure}

We argue that it is necessary to differentiate between different types of unsafe behaviour during training as evaluation as demonstrated in Fig.~\ref{fig:EMCC_DifferentChains}, especially from the perspective of safe exploration.
In Fig~\ref{fig:EMCC_DifferentChains}, despite the two trajectories generated by different policies, but it's clear to see that the right one is more informative as it explores more on the boundary of the infeasible sets and closer to the optimal policy. These transitions on the edge will help the critic and actor to learn a more accurate estimation and prediction, allowing the agent to perform optimization more precisely.
Whereas the left trajectory explores more in the grey area, which is not as valuable as the boundary from the safe exploration perspective as violating the constraints for a long period of time would be more harmful than a few occasional violations. It also helps less for the agent to clearly identify where the safe boundaries are.

With these concerns, we introduce a new metric that quantifies the safe exploration capability of SafeRL algorithms: \textbf{E}xpected \textbf{M}aximum \textbf{C}onsecutive \textbf{C}ost steps (EMCC), which evaluates the severity of unsafe actions based on their consecutive occurrences during training. EMCC is calculated per rollout by taking the maximum of the maximum consecutive cost steps per trajectory divided by the respective trajectory length.  This metric is particularly adept at distinguishing between prolonged and occasional safety violations, providing in-depth understanding of safety during the exploration phase of SafeRL.
We believe that EMCC could help the community for designing SafeRL exploration strategies and brings more insights of different SafeRL algorithms.

Our primary contributions are as follows:
\begin{itemize}
  \item Introduction to EMCC metric, designed to enhance the evaluation of safe exploration within SafeRL frameworks.
  \item Development of a new benchmark task set, Circle2D, tailored for the SafeRL community. This task set features four distinct levels of difficulty and is designed for quick evaluation and easy visualization.
  \item Comprehensive benchmarking of various SafeRL algorithms across different tasks, providing a detailed analysis of their performance in terms of safe exploration.
\end{itemize}

\section{Related work}
\begin{figure*}[t]
    \begin{subfigure}[b]{4.2cm}
        \centering
        \includegraphics[width=\textwidth]{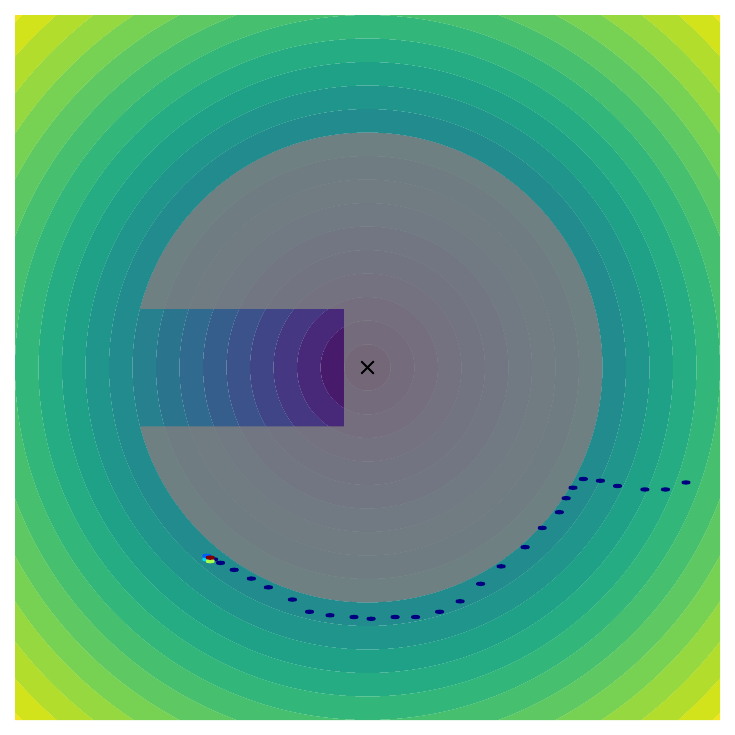} 
        \label{fig:subim1}
    \end{subfigure}\hfill
    \begin{subfigure}[b]{4.2cm}
        \centering
        \includegraphics[width=\textwidth]{
        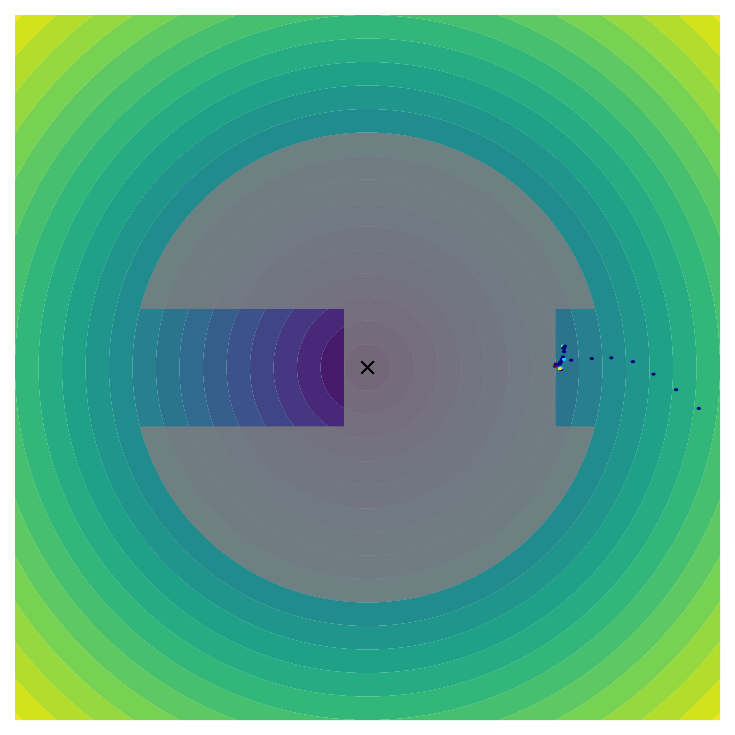}
        \label{fig:subim2}
    \end{subfigure}\hfill
    \begin{subfigure}[b]{4.2cm}
        \centering
        \includegraphics[width=\textwidth]{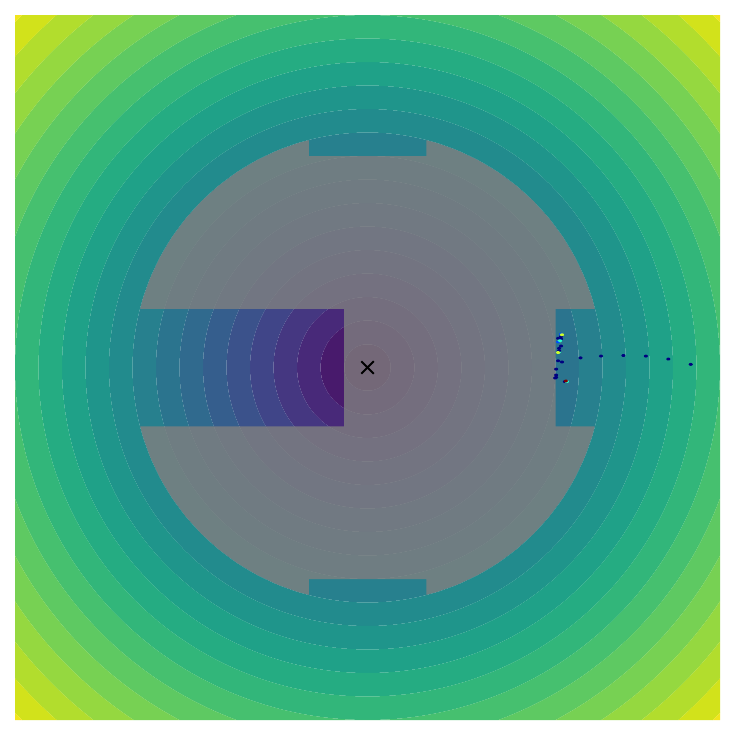}
        \label{fig:subim3}
    \end{subfigure}\hfill
    \caption{Different Circle2D levels. \textbf{Left}: Level 0 and 1, \textbf{Mid}: level 2, \textbf{Right}: level 3. With increasing level more cutouts are added to the cost region which results in more local optima. The dotted line are trajectories with the dot color denoting the frequency of visiting the state. Note that level 0 and 1 have the same cost region structure but level 0 differs with a non-penetrable cost region. The concentric circles in the background visualize the reward at a state with the center X being the global optimum.}
    \label{fig:Circle2DLevels}
    \vspace{15pt}
\end{figure*}

\paragraph{SafeRL Algorithms}
Numerous studies have proposed diverse methods to enhance the safety of Reinforcement Learning (RL). Comprehensive reviews of these approaches can be found in~\citep{DBLP:conf/ijcai/ZhaoHCWL23, DBLP:journals/jmlr/GarciaF15, DBLP:journals/ml/Dulac-ArnoldLML21, DBLP:conf/l4dc/MaLLZC22}.

Safe Policy Search integrates techniques from nonlinear programming into policy gradient methods~\citep{4354030} and builds theoretical frameworks for lifelong RL to ensure safety via gradient projection~\citep{DBLP:conf/icml/Bou-AmmarTE15}. Constrained Policy Optimization (CPO)~\citep{DBLP:conf/icml/AchiamHTA17} emerged as the first general-purpose method employing a trust-region approach with theoretical guarantees. Conditional Value-at-Risk (CVaR) has also been utilized to optimize Lagrangian functions with gradient descent~\citep{DBLP:conf/nips/ChowTMP15, DBLP:journals/jmlr/ChowGJP17}. 

Extensions to Soft Actor-Critic (SAC) incorporate cost functions and employ Lagrange-multipliers to handle constraints, although training robustness issues arise when constraint violations are infrequent~\citep{DBLP:conf/corl/HaXTLT20}. This framework has been used for multitask learning on real robots, with safety ensured by learning predictive models of constraint violations~\citep{DBLP:journals/corr/safety_critic, DBLP:conf/aaai/YangSTS21, DBLP:conf/ijcai/YingZ0Y0022}. Lyapunov functions provide another approach, projecting policy parameters onto feasible solutions during updates, applicable across various policy gradient methods like DDPG or PPO~\citep{DBLP:conf/nips/ChowNDG18, DBLP:journals/corr/abs-1901-10031}. SafeDreamer~\citep{huang2023safedreamer} uses the Dreamer~\citep{DBLP:conf/iclr/HafnerLB020} architecture but also takes safety into consideration


\paragraph{SafeRL Benchmarks}
Several SafeRL benchmarks \citep{Achiam2019BenchmarkingSE}, \citep{ji2023safetygymnasium}, \citep{liu2023datasets}, \citep{zhao2023guard} have been proposed often focusing on different aspects of SafeRL.
In \citep{Achiam2019BenchmarkingSE} Safety Gym is introduced as the first standard set of environments for SafeRL. With Safety-Gymnasium \citep{ji2023safetygymnasium} extends\citep{Achiam2019BenchmarkingSE} with more agents, tasks and benchmarked algorithms. GUARD \citep{zhao2023guard} benchmarks TRPO-based SafeRL algorithms on a broad set of tasks and agents while \citep{liu2023datasets} focus on offline SafeRL.

As metrics for quantifying safety these benchmarks use the (normalized) cost return of the trained policy. For evaluating safety during training next to learning curves \citep{Achiam2019BenchmarkingSE} and \citep{zhao2023guard} provide the metric of cost rate (sum of all costs divided by number of environment interaction steps of the training) for quantifying the general safety of the training process. Compared to the employed metrics of these benchmarks we propose a new metric for quantifying the safe exploration.

\section{Background}
\paragraph{Markov Decision Process}
A Markov Decision Process (MDP) is formalized as a tuple 
$(\{S, A, P, r, \gamma\}$ where $S$ is the state space, $A$ the action space, P the transition model of the environment, $r(s'|s,a)$ the reward function, describing the reward given when transitioning from state $s$ to next state $s'$ with action $a$, and $\gamma \in [0,1]$  the discount factor. The objective to maximize expected discounted cumulative reward, which is defined as: 
\begin{eqnarray}\label{eq:Return}
J_R(\pi) = \mathbb{E}_{\pi}\left[\sum_{t=0}^{T} \gamma^t r(s_{t+1}|s_t, a_t) \right]
\end{eqnarray}
where $\pi$ is defined as the policy which outputs the action distribution given a state.

\paragraph{Constrained Markov Decision Process}
Constrained Markov Decision Processes (CMDPs)~\citep{Altman1999CMDP} extend MDPs to the constrained optimization problem by augmenting the objective with one or multiple cost functions $C_i(s'|s,a)$ in analogy to the reward function, the cost threshold $D_i$ respectively and discount factor \(\gamma_c \in [0,1]\). We define $\mathit{J}_\mathit{C} (\pi)$ as the expected discounted cumulative cost. Then we have the feasible set of policy defined as:
\begin{align}
\Pi_{\mathit{C}} = \{\pi \in \Pi : \mathit{J}_\mathit{C} (\pi) -  \mathit{D} \leq 0\}
\end{align}
The constrained optimization problem can be written as
\begin{align}
\pi^{*} = \arg \max_{\pi \in \Pi_{\mathit{C}}} \mathit{J}(\pi)
\end{align}
which maximizes the return while respects all constraints.

\paragraph{Existing Metrics for SafeRL}
For measuring performance and safety during and after training the following metrics are used in the existing benchmarks \citep{Achiam2019BenchmarkingSE}, \citep{ji2023safetygymnasium}, \citep{liu2023datasets} and \citep{zhao2023guard}.
\begin{itemize}
  \item Average episode return $J_R$
  \item Average episodic sum of costs $J_C$
  \item Cost rate $\rho_c$: sum of all costs divided by number of environment interaction steps during training.
  \item Conditional Value-at-Risk (CVaR): 
For a bounded-mean random variable $Z$, the value-at-risk (VaR) of $Z$ with confidence level $\alpha\in(0,1)$ is defined as:
\begin{equation}
    \label{VaR}
    \mathrm{VaR}_{\alpha}(Z) = F_z^{-1}(1-\alpha),
\end{equation}

where $F_z = P(Z \leq z)$ is the cumulative distribution function (CDF); 
and the conditional value-at-risk (CVaR) of $Z$ with confidence level $\alpha$ is defined as the expectation of the $\alpha$-tail distribution of $Z$ as
\begin{equation}
    \label{CVaR}
    \mathrm{CVaR}_{\alpha}(Z) = \mathbb{E}_{z\sim Z}\{z|z\ge \mathrm{VaR}_{\alpha}(Z)\}.  
\end{equation}
 

\end{itemize}

\section{Circle2D Environment}
For rapid evaluation of safe exploration, we introduce the "Circle2D" environment, which features four levels of difficulty, ranging from 0 to 3 as depicted in Fig.~\ref{fig:Circle2DLevels}. This environment serves as a simplified model of real-world scenarios involving complex cost regions, such as areas exceeding speed limits or zones a cleaning robot must avoid. Although these real-world scenarios present greater complexity, they share the underlying principle of navigating cost regions, which must be strategically avoided. The Circle2D environment, by focusing on the exploration of these cost boundaries, offers an effective and straightforward means for assessing the safe exploration strategies of agents, making it a valuable tool for examining safety as a test environment.

The code of the environment is open-sourced and offers the standard SafetyGymnasium-style~\cite{ji2023safetygymnasium} interface and rich customization choices (See Tab.~\ref{tab:EnvironmentConfigParameter}) for future development.

\paragraph{Circle2D details}

The Circle2D environment features a global infeasible optimum located within a circular cost region, marked by a black X. The reward is based on the normalized distance to this optimum. Agents start in a rectangular area to the right of the cost region, tasked with navigating towards the global feasible optimum, potentially achieving an discounted return between $-11$ and $-12$, which varies by initial conditions and level.

Levels 0 and 1 share the same cost region structure, differing only in interaction responses: Level 0's cost region cannot be penetrated, causing any interaction to revert the action and a step cost of $1$. Levels 1 to 3 allow penetration, increasing the challenge of safe exploration as the agents might focus on gaining more rewards while overlooking the costs. Levels 2 and 3 further complicate navigation by introducing additional cutouts in the cost region, creating local optima. Costs are incurred for both interacting with and penetrating the cost region across all levels, with a maximum episode length of 50 steps.

\section{Expected maximum consecutive steps: EMCC}

Fig~\ref{fig:EMCC_DifferentChains} has depicted two scenarios where conventional metrics cannot well differentiate. To tackle this challenge and measure the safe exploration, we propose a new metric Expected maximum consecutive steps (EMCC), which defines as:
\begin{eqnarray}\label{eq:mcc}
MCC_D = \max_{\tau \in D}(\frac{d_{\tau}^{max}}{l_{\tau}})
\end{eqnarray}
\begin{eqnarray}\label{eq:emcc}
EMCC = \mathbb{E}_{D\sim\pi} \left[MCC_D\right]
\end{eqnarray}
where $D$ is the set of rollouts generated by a policy $\pi$ and $\tau$ is a subset of $D$ which represents a consecutive trajectory with arbitrary length. Note that policy $\pi$ changes between rollouts due to the online update during the training. In the case of Fig~\ref{fig:EMCC_DifferentChains}, the MCC without normalization by the total episode length of the left trajectories will be $8$ and the right one will be $3$. EMCC considers multiple rollouts during a period of training time to give an average estimation.

In its general form, Eq.~\ref{eq:emcc} calculates one value for the whole training process. To capture the changing behaviour in exploration during training, we divide the training process into three parts and calculate EMCC per part. This allows more precise interpretation for safe exploration as later rollouts cannot influence the EMCC value of the first third of the training process. Making this distinction is especially relevant for the first third as the most exploration is expected at the beginning. We denote EMCC split into the different training parts uniformly as $EMCC_\beta$ with $\beta$ showing the relevant training part. $EMCC_{0.33}$ combines data from the start to $33\%$ of the training, $EMCC_{0.66}$ for $33\%$ to $66\%$ and $EMCC_{0.99}$ for $66\%$ to $99\%$.

Furthermore we augment \(EMCC_\beta\) with the conditional Value-at-Risk (CVaR) \citep{Rockafellar2000CVAR} to focus on the highest MCC values of the MCC distribution per training part . As we associate the most prolonged safety violations with the most risky behaviour augmenting with CVaR further enhances \(EMCC_\beta\) as a safety measure. We follow the definition and notation of \cite{Yang2021WCSAC} for using a positive scalar \(\alpha \in [0,1)\) as risk level in safety and apply it to \(EMCC_\beta\):
\begin{eqnarray}\label{eq:emcc_beta_cvar}
EMCC_\beta^\alpha = \mathbb{E}_{D_\beta\sim\pi} \left[ MCC_{D_\beta} | MCC_{D_\beta} \geq F_{MCC_\beta}^{-1}(1-\alpha) \right]
\end{eqnarray}

In Eq.~\ref{eq:emcc_beta_cvar} \(D_\beta\) denotes rollouts of the training part associated with \(\beta\) and \(F_{MCC_\beta}^{-1}(1-\alpha)\) is the \(\alpha\)-percentile with \(F_{MCC_\beta}\) being the cumulative distribution function of the distribution of corresponding \(MCC\) values.

\begin{figure}[h]
    \begin{subfigure}[b]{2.5cm}
        \centering
        \includegraphics[width=\textwidth]{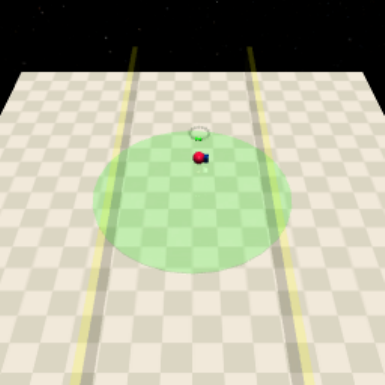} 
        \label{fig:circle}
    \end{subfigure}\hfill
    \begin{subfigure}[b]{2.5cm}
        \centering
        \includegraphics[width=\textwidth]{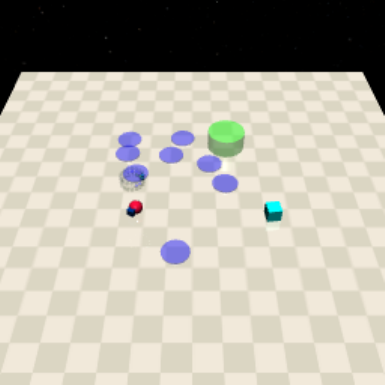}
        \label{fig:goal}
    \end{subfigure}\hfill
    \begin{subfigure}[b]{2.5cm}
        \centering
        \includegraphics[width=\textwidth]{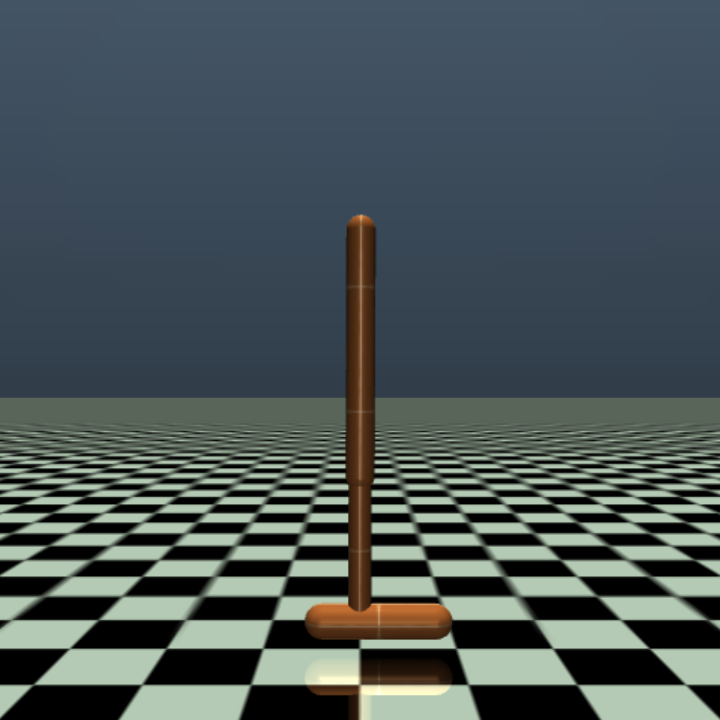}
        \label{fig:hopper}
    \end{subfigure}\hfill
    \caption{\textbf{Left}: SafetyPointCircle1-v0, \textbf{Mid}: SafetyPointGoal1-v0, \textbf{Right}: SafetyHopperVelocity-v1}
    \label{fig:SafetyGymnasiumTasks}
    \vspace{15pt}
\end{figure}

\begingroup
\begin{table}[h]
\caption{Hyperparameters of on- and off-policy algorithms for Circle2D tasks}
\label{tab:HP_Circle2D}
\centering
\renewcommand{\arraystretch}{1.1}
\resizebox{\columnwidth }{!}
{
\begin{tabular}{l|l|l|l|l|l} 
\toprule
Circle2D tasks & TRPO-Lag & CPO  & SAC-Lag  & SAC-LB & WCSAC$_{0.5}$\\
\hline
episodes per epoch & 10 & 10 & 1 & 1 & 1 \\
total timesteps & 1e6 & 1e6& 1e6& 1e6& 1e6\\
batch size & 256 & 256 & 256 & 256 & 256 \\
discount factor & 0.99&0.99 &0.99 & 0.99&0.99 \\
learning rate &  3e-4 &  3e-4 &3e-4 &3e-4 & 3e-4\\
critic update iterations & 10 & 10 & / & / & /\\

GAE $\lambda$ &0.95 &0.95 & /&/ & /\\
conjugate gradient iterations & 15& 15 &/ &/ & /\\
linesearch max steps & 15 & 15 & / & / & /\\
target kl-divergence & 0.01 & 0.01 & / & / & / \\
initial Lagrange multiplier & 0.001 & / & 0 & / & /\\
Lagrange multiplier learning rate & 0.005 & / & 3e-4 & / & /\\

random steps & / & / & 5000 & 5000 & 5000\\
gradient steps per epoch & / & / & 5& 5& 5\\
replay buffer size & / & / &50000& 50000 & 50000\\
log barrier factor & / & / & 3 & / & / \\

\bottomrule
\end{tabular}
}
\end{table}
\endgroup

\begingroup
\begin{table}[h]
\caption{Hyperparameters of on- and off-policy algorithms for Safety-Gymnasium tasks that are different from Circle2D tasks}
\label{tab:HP_SafetyGymnasium}
\centering
\renewcommand{\arraystretch}{1.2}
\resizebox{\columnwidth }{!}{
\begin{tabular}{l|l|l|l|l|l} 
\toprule
Safety-Gymnasium tasks & TRPO-Lag & CPO  & SAC-Lag  & SAC-LB & WCSAC$_{0.5}$\\
\hline
episodes per epoch & 10 (20 SafetyCircle) & 10 (20 SafetyCircle) & 1 & 1 & 1 \\
total timesteps & 1e7 & 1e7& 1e7& 1e7& 1e7\\
random steps & / & / & 50000 & 50000 & 50000\\
replay buffer size & / & / &500000& 500000 & 500000\\

\bottomrule
\end{tabular}
}
\end{table}
\endgroup

To clarify EMCC calculation we provide a conceptual calculation process in 5 steps: \\
\textbf{1. Initialization \(\beta\):} Define which part of the training process to evaluate with EMCC. \\
\textbf{2. Per trajectory calculation:} For each rollout associated with training part \(\beta\) calculate for each trajectory the maximum number of consecutive cost steps \(d_{\tau}^{max}\) and normalize with the respective trajectory length. \\
\textbf{3. MCC per rollout:} For each rollout find the maximum of the corresponding normalized numbers of consecutive cost steps (result from step 2). This is the MCC value of the corresponding rollout. \\
\textbf{4. Risk level \(\alpha\):} From the distribution of MCC values only keep the share of highest values defined by risk level \(\alpha\). \\
\textbf{5. EMCC for training part \(\beta\):} Average over the remaining MCC values to get \(EMCC_\beta^\alpha\) for training part \(\beta\) and risk level \(\alpha\).

\section{Experiments}
\begingroup
\begin{table}[h]
\caption{Circle2D tasks with cost limit 5. EMCC and cost rate \(\rho_c\) for quantifying safety during training and return \(J_R\), cost return \(J_C\) and \(CVaR_{0.5}\) cost return averaged from 100 episodes after training. All metrics are averaged over 3 seeds. The best (lowest) EMCC values and cost rate are highlighted. For \(J_R\), \(J_C\) and \(CVaR_{0.5}\) the algorithms are highlighted that score the highest return \(J_R\) while adhering to the cost limit.}
\label{tab:Circle2DResults}
\centering
\renewcommand{\arraystretch}{1.2}
\resizebox{\columnwidth }{!}{
\begin{tabular}{l|l|l|l|l|l} 
\toprule
Circle2D-0 & TRPO-Lag & CPO  & SAC-Lag  & SAC-LB & WCSAC$_{0.5}$\\
\hline
\(EMCC^{0.1}_{0.33}\)    & 0.68   & 0.56  & 0.61   &  \textbf{0.48} & 0.51\\
\(EMCC^{0.1}_{0.66}\)    &  0.54  &  0.56 &  0.52  &  \textbf{0.41} & \textbf{0.41}\\
\(EMCC^{0.1}_{0.99}\)   &  0.69  &  0.63 &  \textbf{0.35}  &  0.40 & 0.48\\
\(\rho_c\)               &  0.11  &  0.09 &  0.11  &  0.09 & \textbf{0.08}\\
\(J_R\)             &  -21.32  & -21.65  & \textbf{-14.47}  &  -15.76 & -19.04\\
\(J_C\)             &  11.61  & 19.58  &  \textbf{0}  & 0 & 1.65\\
\(CVaR_{0.5}\)  &  23.23  & 28.79  & \textbf{0} & 0  & 3.29\\
\bottomrule

Circle2D-1 & TRPO-Lag & CPO  & SAC-Lag  & SAC-LB & WCSAC$_{0.5}$\\
\hline
\(EMCC^{0.1}_{0.33}\)    & \textbf{0.46}   & 0.49  & 0.62   &  0.47 & 0.53\\
\(EMCC^{0.1}_{0.66}\)    & \textbf{0.22}   & 0.44  & 0.63   &  0.28 & 0.37\\
\(EMCC^{0.1}_{0.99}\)    & \textbf{0.22}   & 0.35  & 0.43  & 0.32 & 0.25\\
\(\rho_c\)               & 0.24   & 0.12  & 0.2  & 0.11 & \textbf{0.07}\\
\(J_R\)          & 4.68  & -25.28  & \textbf{-13.7}   & -15.56 & -18.09 \\
\(J_C\)          & 9.39  &  6.98 & \textbf{3.32}  & 0.50 & \textbf{0.33} \\
\(CVaR_{0.5}\) & 9.95  & 11.87  &  \textbf{3.64}  & 0.67 & \textbf{0.66}\\
\bottomrule

Circle2D-2 & TRPO-Lag & CPO  & SAC-Lag  & SAC-LB & WCSAC$_{0.5}$\\
\hline
\(EMCC^{0.1}_{0.33}\)    & \textbf{0.42}   & 0.45 & 0.6  & 0.45 & 0.56 \\
\(EMCC^{0.1}_{0.66}\)    &  \textbf{0.18}  & 0.37 & 0.52  & 0.57  & 0.44\\
\(EMCC^{0.1}_{0.99}\)    &  \textbf{0.18}  &  0.30 & 0.51 & 0.57 & 0.49\\
\(\rho_c\)               &  0.21  &  0.12 & 0.20  & \textbf{0.11} & 0.16\\
\(J_R\)          & -4.23  & -22.29 &  -7.53 & -14.15 & \textbf{-11.56} \\
\(J_C\)          & 9.00  & 4.94 & 6.32  & 0 & \textbf{3.49} \\
\(CVaR_{0.5}\) & 9.00  & 6.35  &  6.61  & 0 & \textbf{3.64}\\
\bottomrule

Circle2D-3 & TRPO-Lag & CPO  & SAC-Lag  & SAC-LB & WCSAC$_{0.5}$\\
\hline
\(EMCC^{0.1}_{0.33}\)    &  \textbf{0.42}  & 0.47 & 0.59   & 0.45  & 0.56\\
\(EMCC^{0.1}_{0.66}\)    &  \textbf{0.19}  &  0.31 &  0.37  &  0.57 & 0.57\\
\(EMCC^{0.1}_{0.99}\)    &  \textbf{0.19} &  0.20 &  0.47 &  0.56 & 0.50\\
\(\rho_c\)               &  0.21  &  \textbf{0.11} &  0.19 &  \textbf{0.11} & \textbf{0.11}\\
\(J_R\)          & -4.41  & -19.22 & -14.10  & \textbf{-14.02} & -15.40 \\
\(J_C\)          & 9.00  & 8.28 & 6.29  & \textbf{0} & 0.01 \\
\(CVaR_{0.5}\) &  9.00 & 12.00  &  6.59  & \textbf{0} & 0.01\\
\bottomrule
\end{tabular}
}
\end{table}
\endgroup

\begingroup
\begin{table}[h]
\caption{Safety-Gymnasium tasks with cost limit 25.0. EMCC and cost rate \(\rho_c\) for quantifying safety during training and return \(J_R\), cost return \(J_C\) and \(CVaR_{0.5}\) cost return averaged from 10 episodes after training. All metrics (training and evaluation) are averaged over 3 seeds.}
\label{tab:SafetyGymnasiumResults}
\centering
\renewcommand{\arraystretch}{1.2}
\resizebox{\columnwidth }{!}{
\begin{tabular}{l|l|l|l|l|l} 
\toprule
SafetyPointCircle1-v0 & TRPO-Lag & CPO  & SAC-Lag  & SAC-LB & WCSAC$_{0.5}$\\
\hline
\(EMCC^{0.1}_{0.33}\)    &  0.83  & \textbf{0.57}  &  0.71  &  0.70 & 0.84\\
\(EMCC^{0.1}_{0.66}\)    &  0.77  & 0.35  &   \textbf{0.26} &  0.27 & 0.56\\
\(EMCC^{0.1}_{0.99}\)   &  0.58  & 0.29  &  \textbf{0.21}  &  0.48 & 0.64\\
\(\rho_c\)               &  0.17  &  \textbf{0.05} &  0.33  &  0.24 & 0.18\\
\(J_R\)             &  19.78  & \textbf{35.52}  &  57.54  &  44.48 & 1.26\\
\(J_C\)             &  74.73  & \textbf{13.23}  &  182.4  &  119.53 & 92.67\\
\(CVaR_{0.5}\)  &  140.13  & \textbf{24.47}  &  202.33  & 137.80  & 139.60\\
\bottomrule

SafetyPointGoal1-v0 & TRPO-Lag & CPO  & SAC-Lag  & SAC-LB & WCSAC$_{0.5}$\\
\hline
\(EMCC^{0.1}_{0.33}\)    & 0.34   & 0.20  & \textbf{0.11}   &  \textbf{0.11} & 0.47\\
\(EMCC^{0.1}_{0.66}\)    &  0.34  & 0.19  &  \textbf{0.07}  & \textbf{0.07}  & 0.48\\
\(EMCC^{0.1}_{0.99}\)   &  0.39  &  0.19 &   \textbf{0.10} &  0.12 & 0.34\\
\(\rho_c\)               &  0.05  & \textbf{0.01}  &  0.06  &  0.06 & 0.16\\
\(J_R\)             &  -1.57  &  -0.36 &  25.73  & 26.41  & -5.09\\
\(J_C\)             &  27.07  & 42.6  &  57.07  & 64,6  & 42.03\\
\(CVaR_{0.5}\)  &  49.00  &  85.20 &  83.00  &  96.93 & 73.13\\
\bottomrule

SafetyHopperVelocity-v1 & TRPO-Lag & CPO  & SAC-Lag  & SAC-LB & WCSAC$_{0.5}$\\
\hline
\(EMCC^{0.1}_{0.33}\)    &  0.85  & 0.79  &  \textbf{0.61}  &  0.76 & 0.69\\
\(EMCC^{0.1}_{0.66}\)    &  0.46  & \textbf{0.31}  &  0.94  &  0.84 & 0.42\\
\(EMCC^{0.1}_{0.99}\)   &   0.24 &  \textbf{0.18} &  0.97  &  0.90 & 0.54\\
\(\rho_c\)               &  0.17  &  \textbf{0.12}  &  0.72  &  0.73 & 0.19\\
\(J_R\)             &  \textbf{867.67}  &  531.26 &  1217.52  &  1174.9 & 458.04\\
\(J_C\)             &  \textbf{15.1}  & 18.73  &  264.93  &  369.33 & 74.60\\
\(CVaR_{0.5}\)  &  \textbf{15.2}  & 19.23  & 311.53   & 447.60  & 75.73\\
\bottomrule
\end{tabular}
}
\end{table}
\endgroup

We use our proposed Circle2D environment with its 4 levels and the Safety-Gymnasium \citep{ji2023safetygymnasium} tasks \textit{SafetyPointCircle1}, \textit{SafetyPointGoal1} and \textit{SafetyHopperVelocity} depicted in Fig.\ref{fig:SafetyGymnasiumTasks} to evaluate the safe exploration process during training. We use a 3 layer MLP with two hidden layers of size 64 and Tanh activation function. Hyperparameters for the Circle2D and Safety-Gymnasium tasks are disclosed in Tab.~\ref{tab:HP_Circle2D} and Tab.~\ref{tab:HP_SafetyGymnasium} respectively.

\paragraph{Algorithms}
We choose algorithms as representatives of their respective classes.

\textbf{TRPO-Lag}: on-policy SafeRL algorithm which extends TRPO as policy search algorithm for CMDPs via introducing a Lagrangian multiplier. Then it solves the results unconstrained optimization problem with dual gradient descent.

\textbf{CPO}: on-policy method~\citep{DBLP:conf/icml/AchiamHTA17} which uses second order method to enforce the constraints during the policy search.

\textbf{SAC-Lag}: off-policy algorithm~\citep{Sehoon2021SACLAG} based on SAC~\citep{Haarnoja2018SAC} and Lagrangian method

\textbf{SAC-LB}: off-policy algorithm~\citep{zhang2024SACLB} based on SAC~\citep{Haarnoja2018SAC} and introduces a linear smoothed log barrier function to replace the Lagrangian multiplier to stabilize the training.

\textbf{WCSAC}: off-policy algorithm~\citep{Yang2021WCSAC} which further extends SAC-Lag by replacing the expected cost return of SAC-Lag with the conditional Value-at-Risk (CVaR) given a risk level. In our experiments we use a risk level of 0.5 as it shows best overall performance in the original work.

\begin{figure*}[h]
    \begin{subfigure}[b]{4.2cm}
        \centering
        \includegraphics[width=\textwidth]{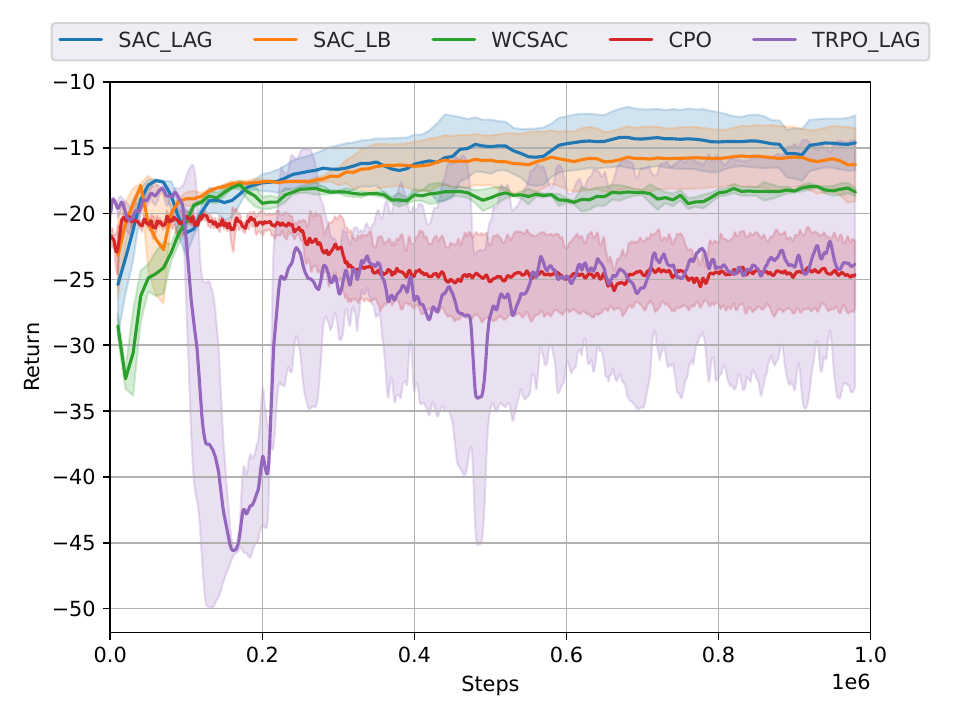}
        \label{fig:Circle2D-0Return}
    \end{subfigure}\hfill
    \begin{subfigure}[b]{4.2cm}
        \centering
        \includegraphics[width=\textwidth]{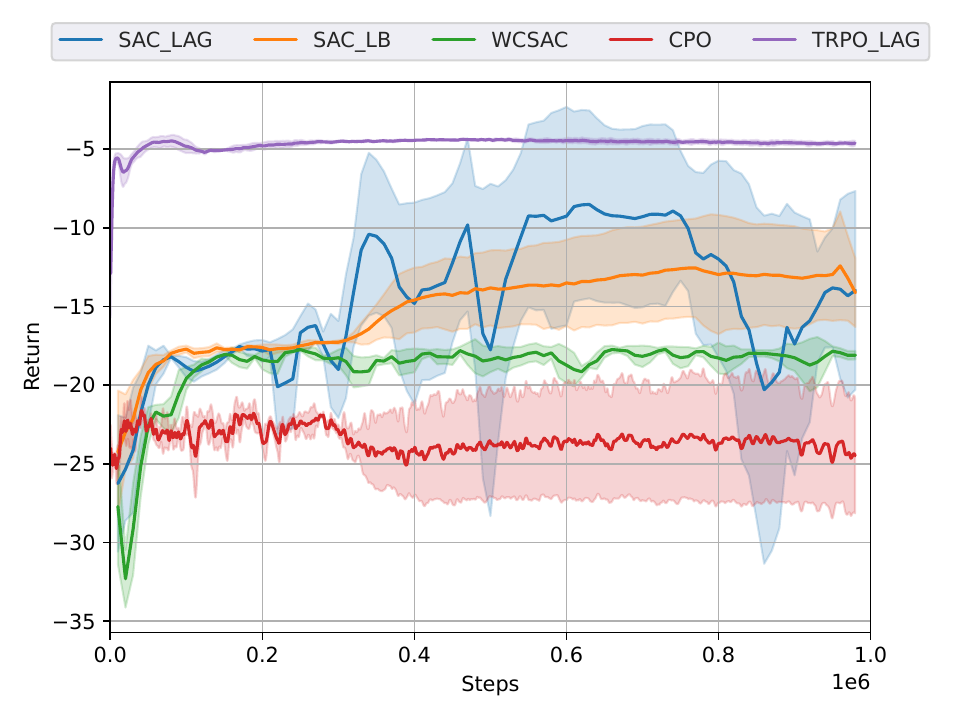}
        \label{fig:Circle2D-1Return}
    \end{subfigure}\hfill
    \begin{subfigure}[b]{4.2cm}
        \centering
        \includegraphics[width=\textwidth]{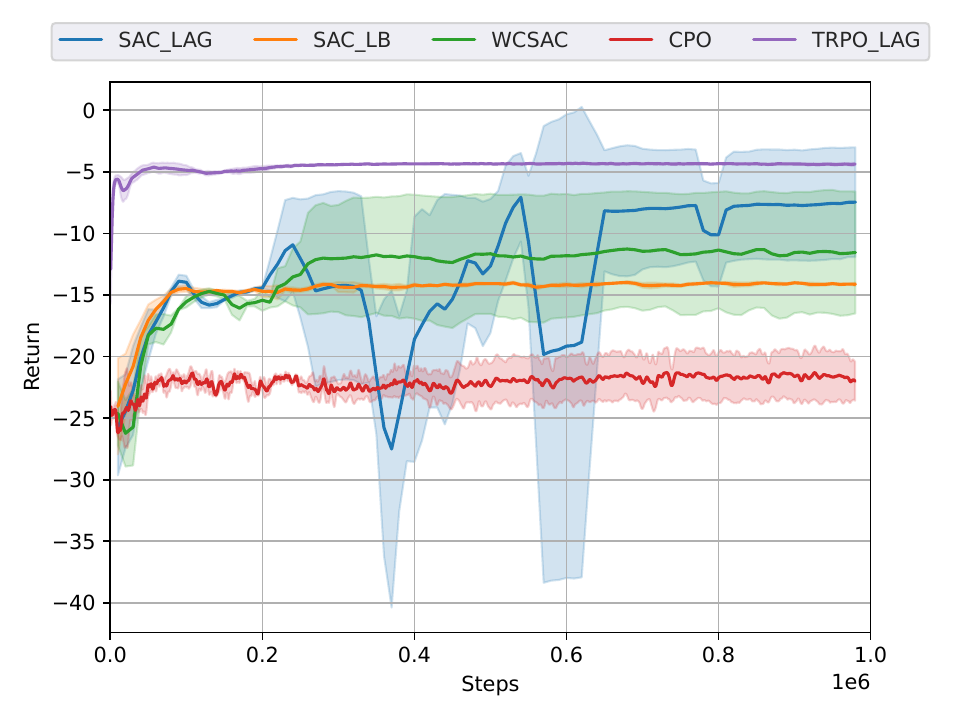}
        \label{fig:Circle2D-2Return}
    \end{subfigure}\hfill
    \begin{subfigure}[b]{4.2cm}
        \centering
        \includegraphics[width=\textwidth]{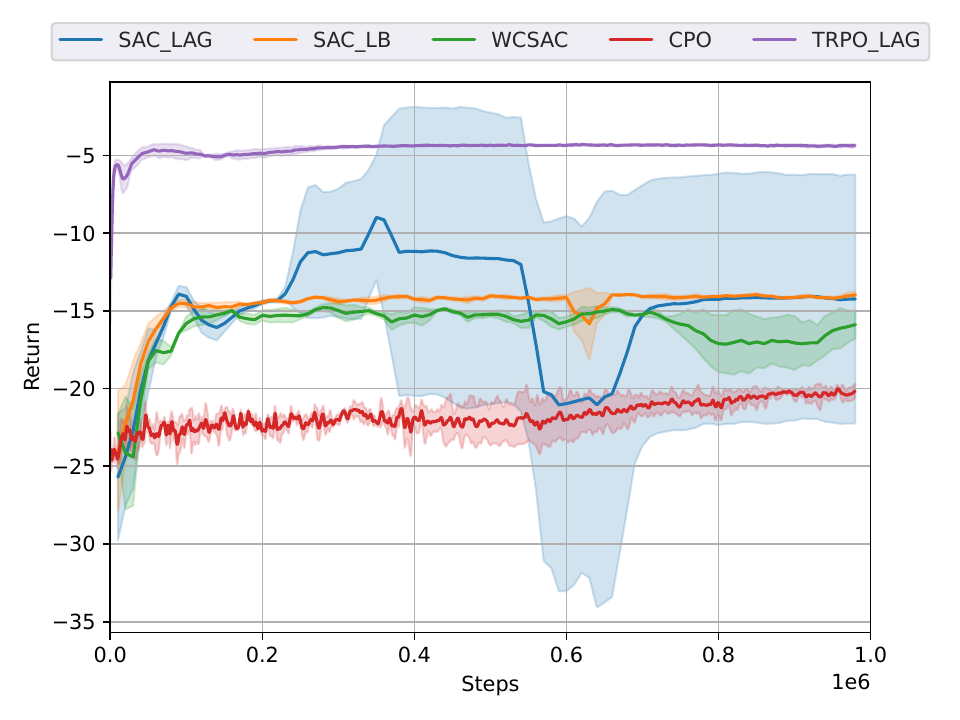}
        \label{fig:Circle2D-3Return}
    \end{subfigure}\hfill

    \begin{subfigure}[b]{4.2cm}
        \centering
        \includegraphics[width=\textwidth]{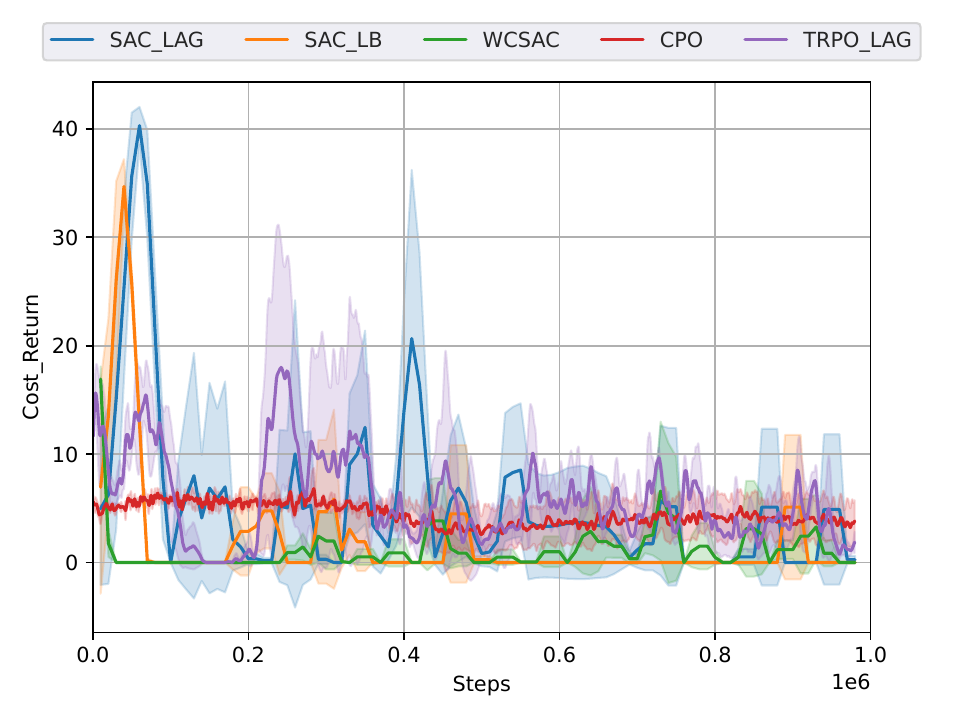}
        \caption{Circle2D-0}
        \label{fig:Circle2D-0CostReturn}
    \end{subfigure}\hfill
    \begin{subfigure}[b]{4.2cm}
        \centering
        \includegraphics[width=\textwidth]{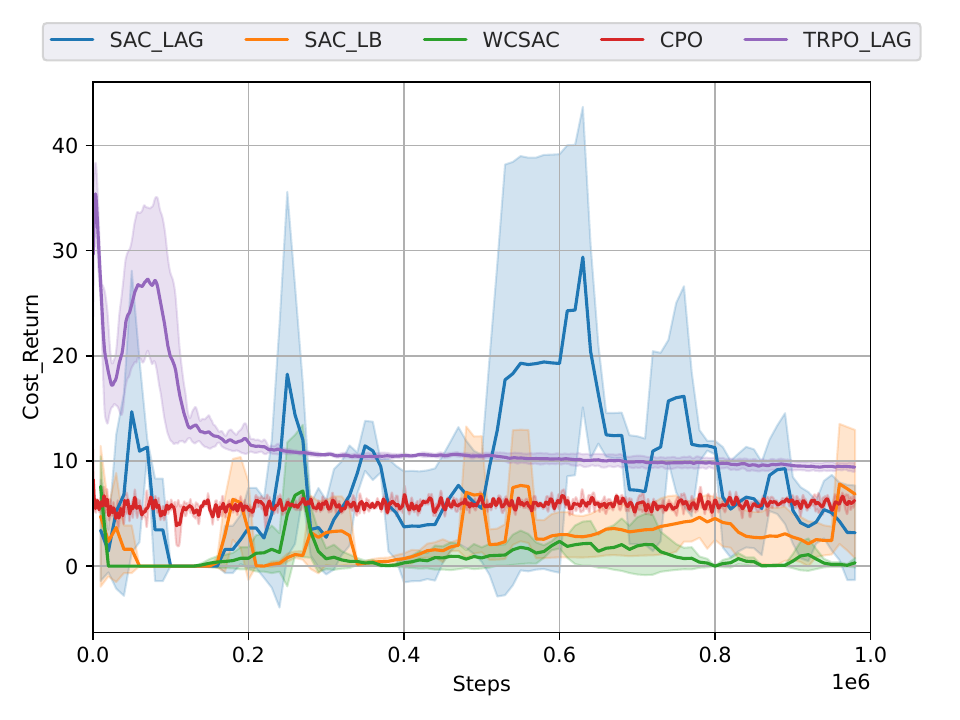}
        \caption{Circle2D-1}
        \label{fig:Circle2D-1CostReturn}
    \end{subfigure}\hfill
    \begin{subfigure}[b]{4.2cm}
        \centering
        \includegraphics[width=\textwidth]{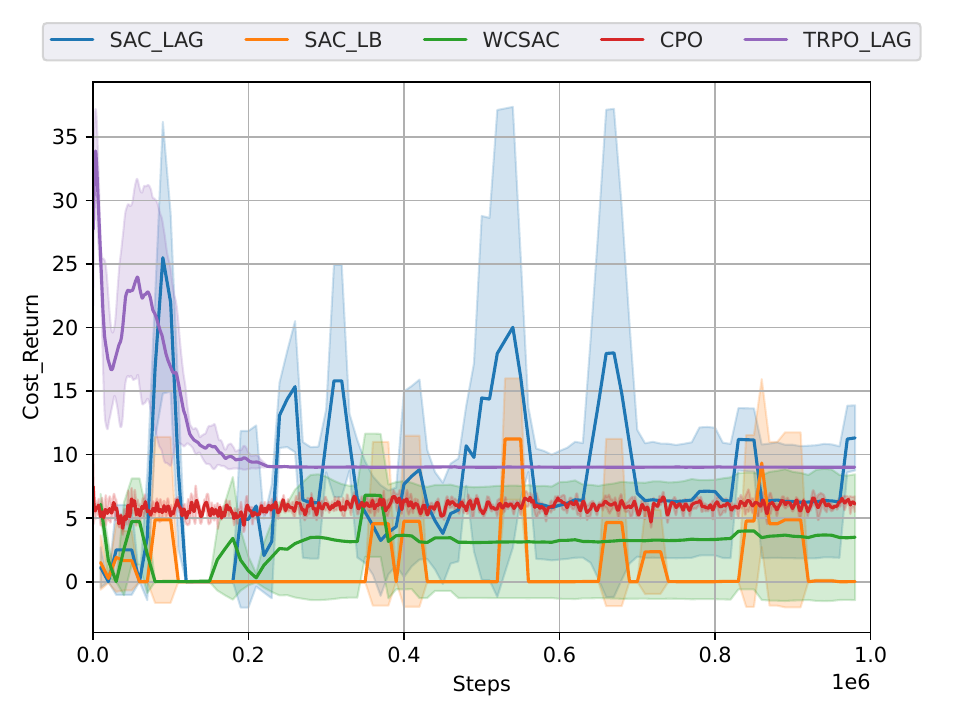}
        \caption{Circle2D-2}
        \label{fig:Circle2D-2CostReturn}
    \end{subfigure}\hfill
    \begin{subfigure}[b]{4.2cm}
        \centering
        \includegraphics[width=\textwidth]{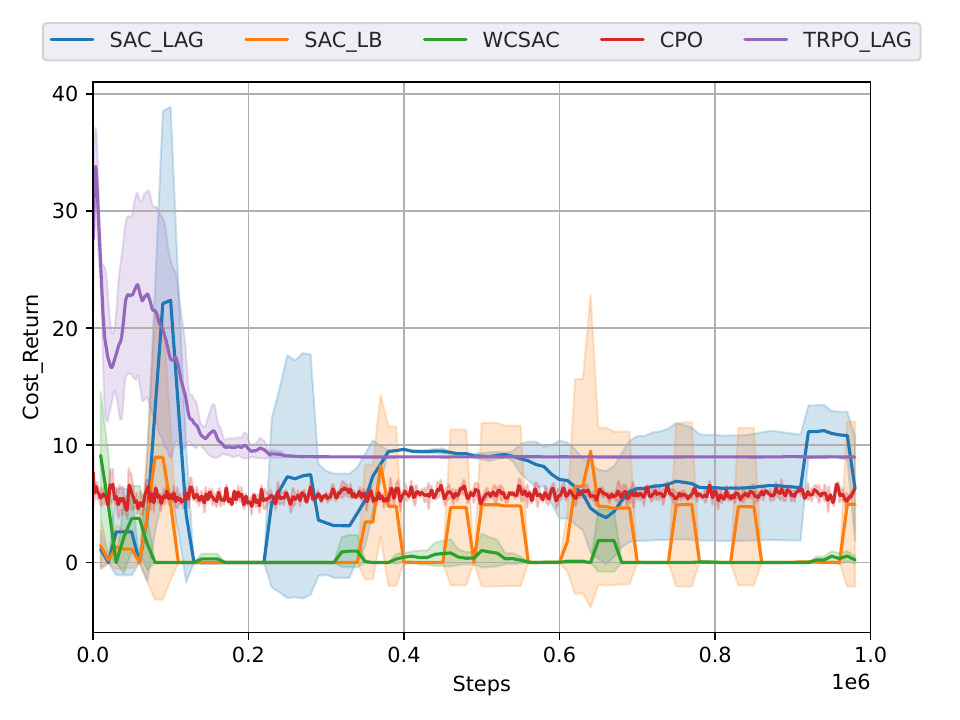}
        \caption{Circle2D-3}
        \label{fig:Circle2D-3CostReturn}
    \end{subfigure}\\
    \vspace{15pt}
    \caption{Training curves for the Circle2D tasks. The curves show the mean and the faint areas the standard deviation of return and cost return of the training process averaged over 3 seeds.}
    \label{fig:Circle2DTrainingCurves}
\end{figure*}

\begin{figure*}[h]
    \begin{subfigure}[b]{5cm}
        \centering
        \includegraphics[width=\textwidth]{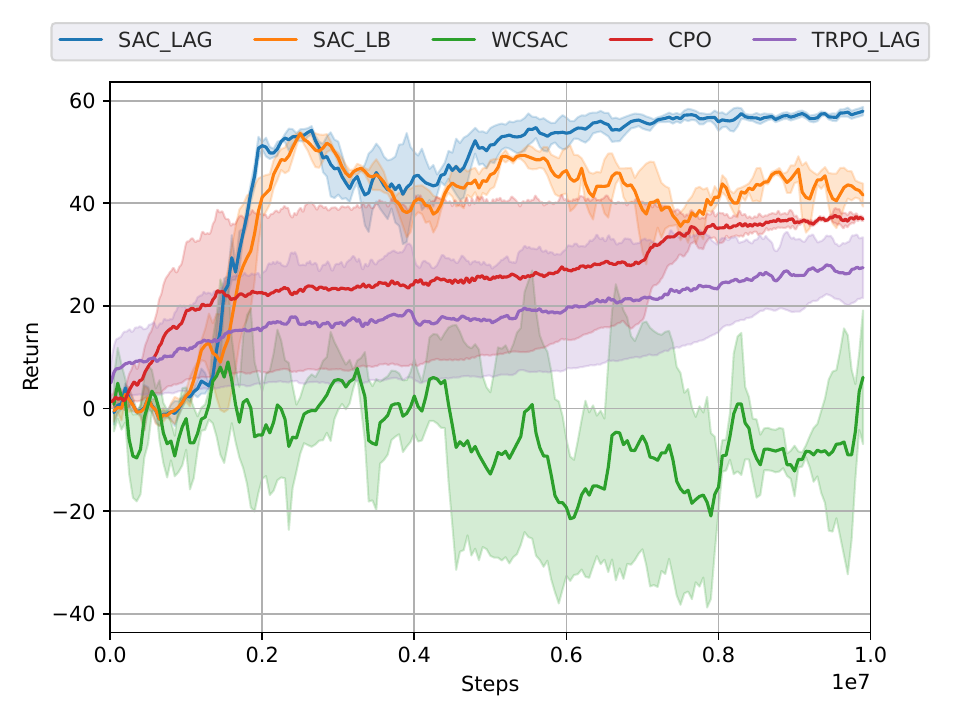}
        \label{fig:SafetyPointCircleReturn}
    \end{subfigure}\hfill
    \begin{subfigure}[b]{5cm}
        \centering
        \includegraphics[width=\textwidth]{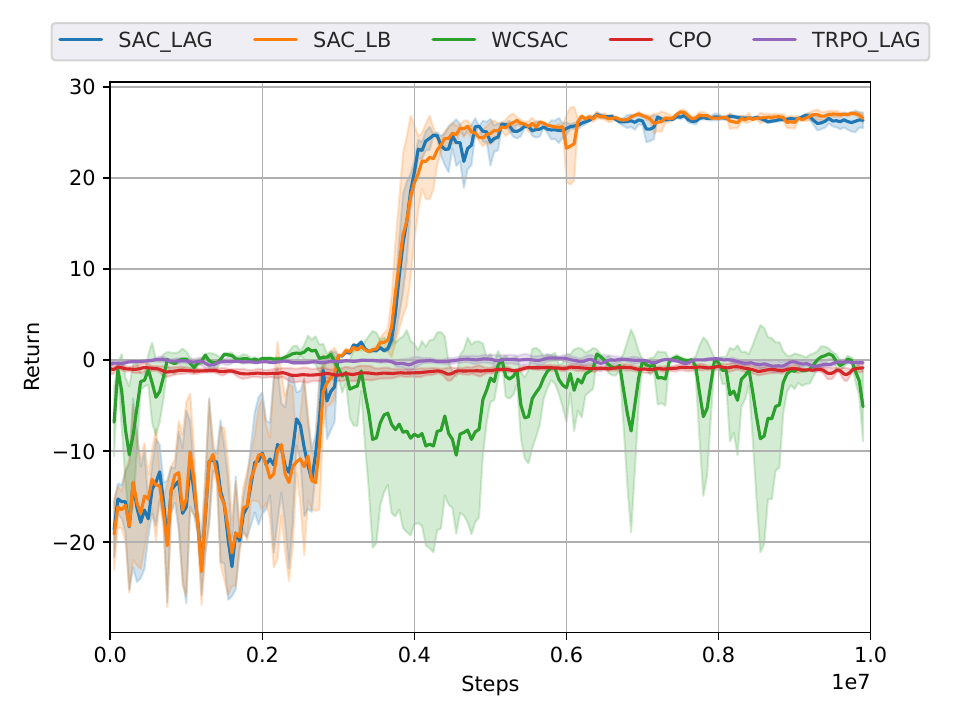}
        \label{fig:SafetyPointGoalReturn}
    \end{subfigure}\hfill
    \begin{subfigure}[b]{5cm}
        \centering
        \includegraphics[width=\textwidth]{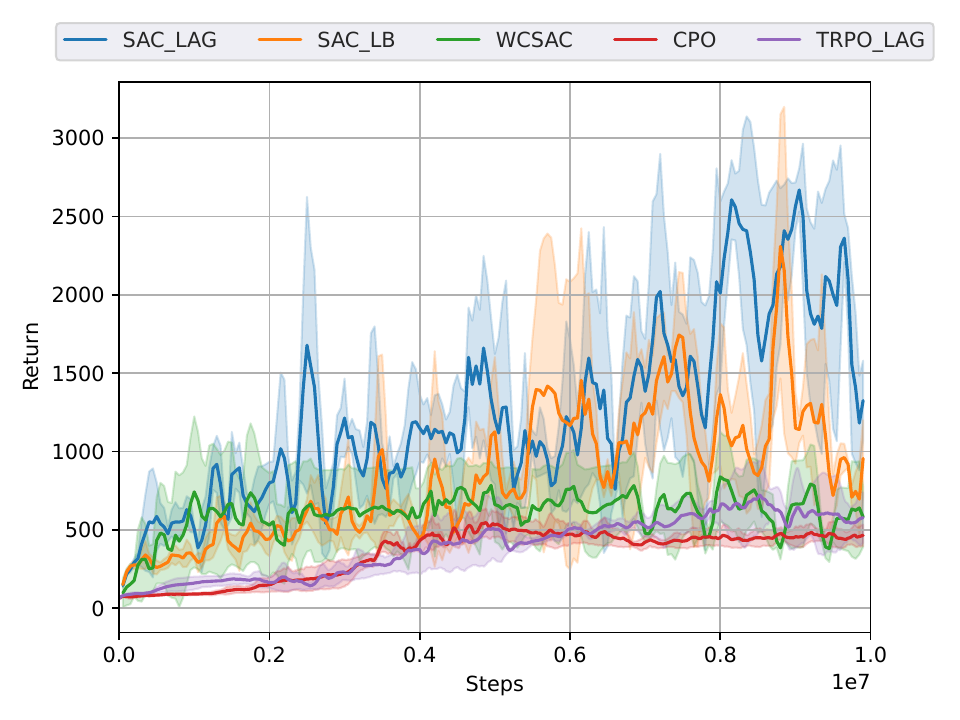}
        \label{fig:SafetyHopperVelocityReturn}
    \end{subfigure}\hfill
    \begin{subfigure}[b]{5cm}
        \centering
        \includegraphics[width=\textwidth]{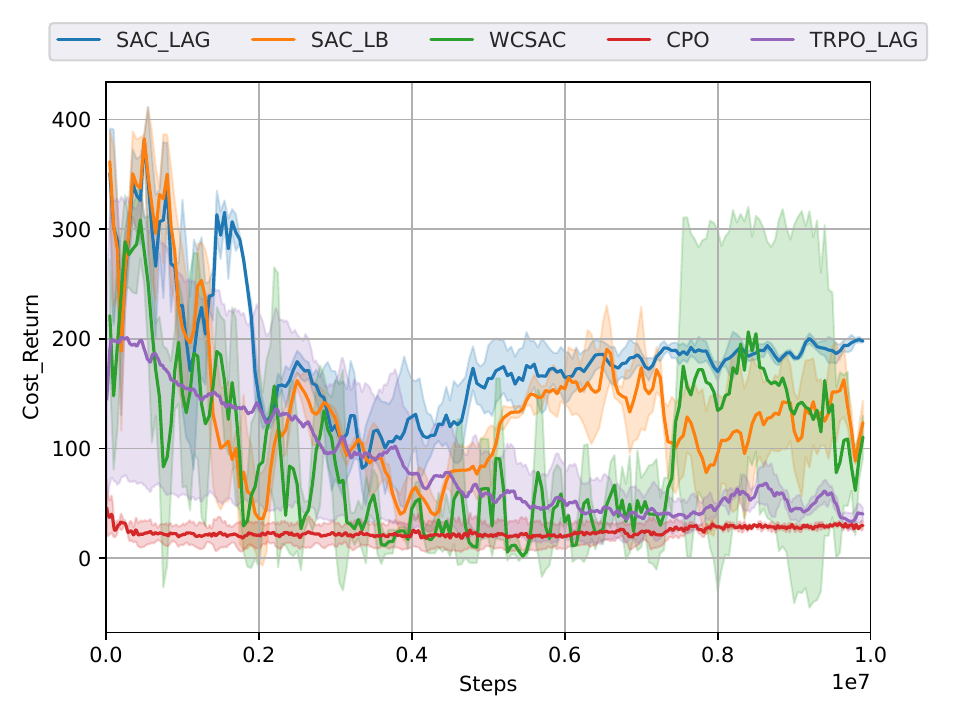}
        \caption{SafetyPointCircle1-v0}
        \label{fig:SafetyPointCircleCostReturn}
    \end{subfigure}\hfill
    \begin{subfigure}[b]{5cm}
        \centering
        \includegraphics[width=\textwidth]{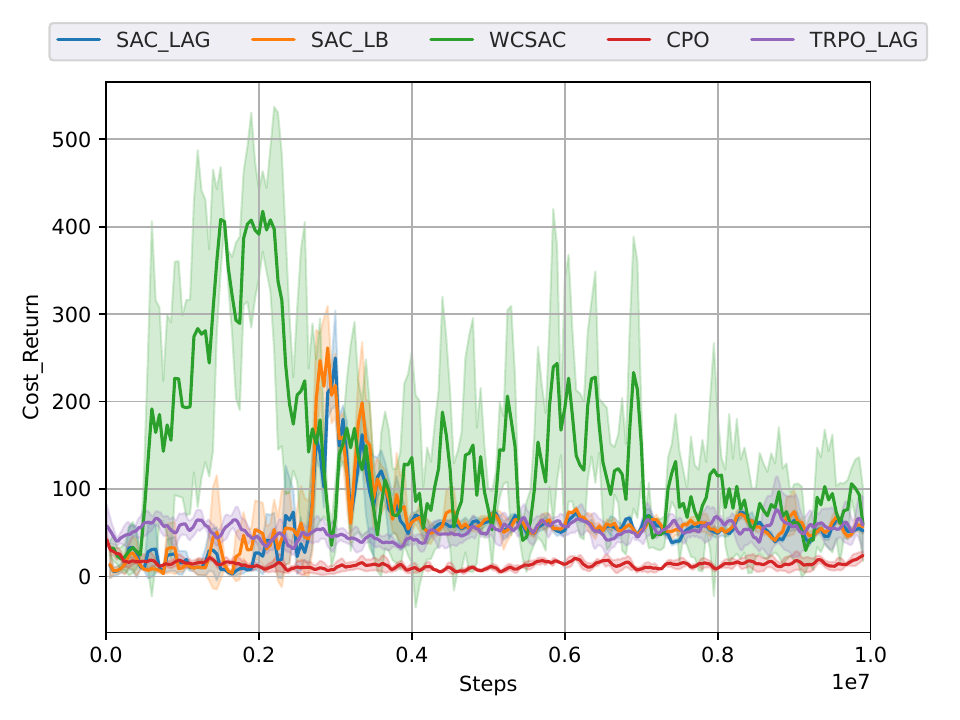}
        \caption{SafetyPointGoal1-v0}
        \label{fig:SafetyPointGoalCostReturn}
    \end{subfigure}\hfill
    \begin{subfigure}[b]{5cm}
        \centering
        \includegraphics[width=\textwidth]{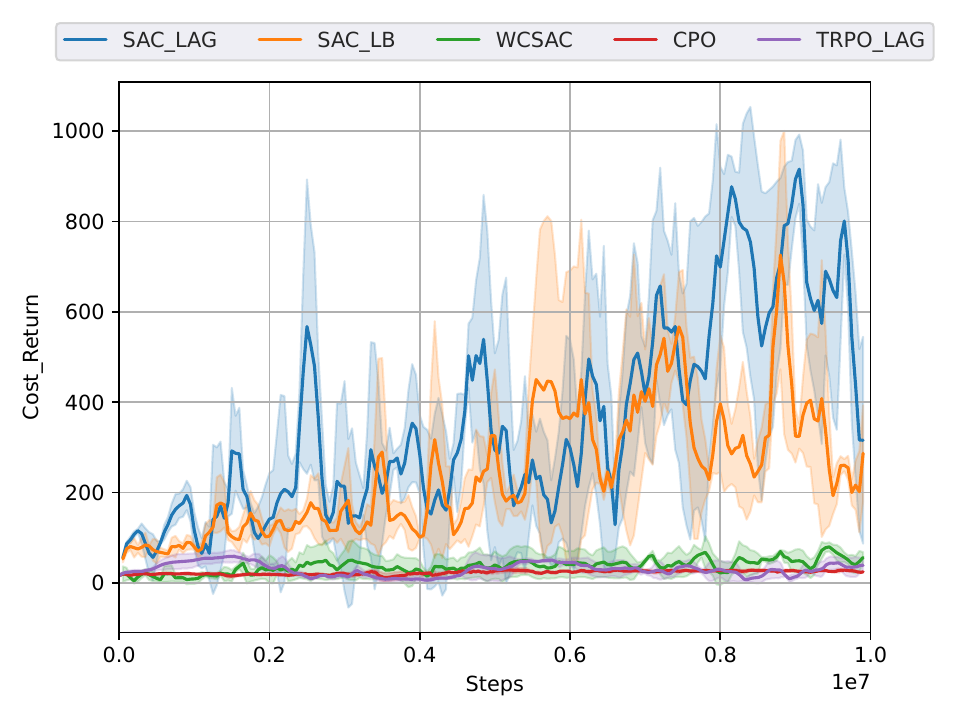}
        \caption{SafetyHopperVelocity-v1}
        \label{fig:SafetyHopperVelocityCostReturn}
    \end{subfigure}\\
    \vspace{18pt}
    \caption{Training curves for the Safety-Gymnasium tasks. The curves show the mean and the standard deviation (faint area) of return and cost return of the training process averaged over 3 seeds.}
    \label{fig:SafetyGymnasiumTrainingCurves}
\end{figure*}

\begin{figure*}[t]
    \begin{subfigure}[b]{3cm}
        \centering
        \includegraphics[width=\textwidth]{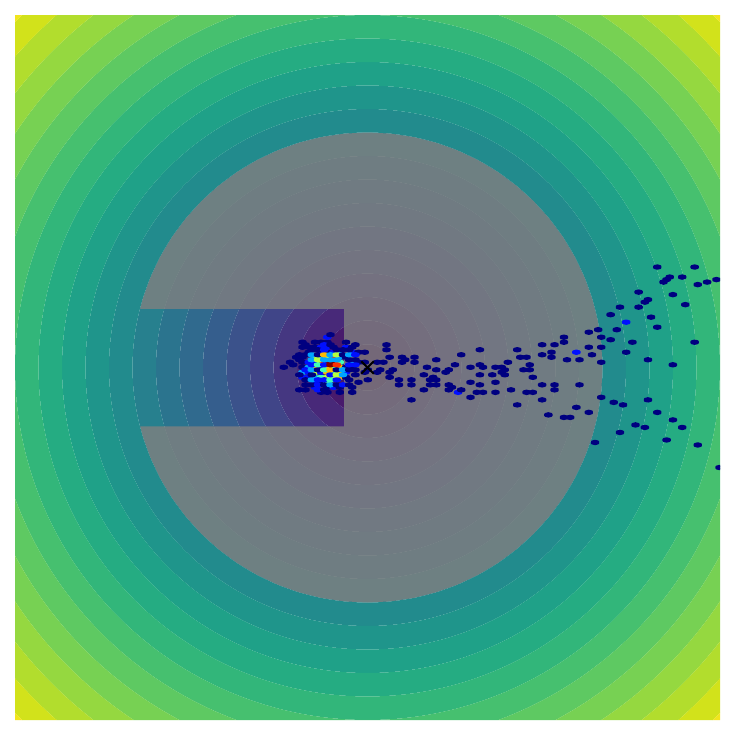}
        \label{fig:TRPOLagfirstThird}
    \end{subfigure}\hfill
    \begin{subfigure}[b]{2.97cm}
        \centering
        \includegraphics[width=\textwidth]{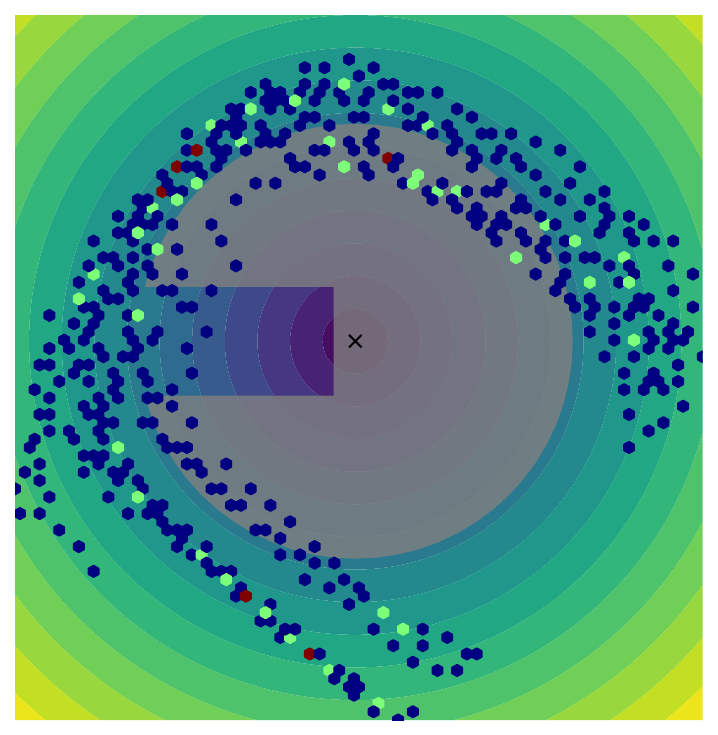}
        \label{fig:CPOfirstThird}
    \end{subfigure}\hfill
    \begin{subfigure}[b]{3cm}
        \centering
        \includegraphics[width=\textwidth]{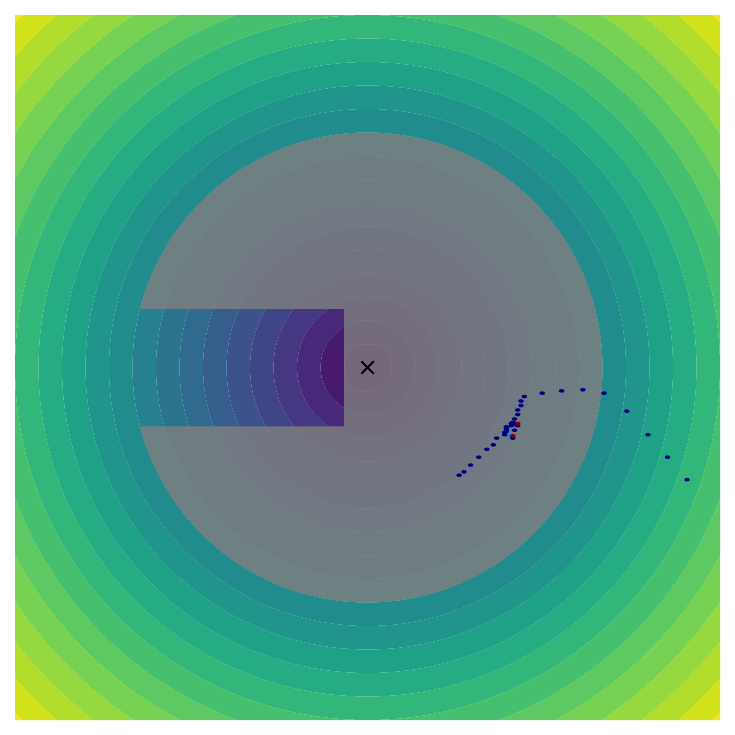}
        \label{fig:SACLagfirstThird}
    \end{subfigure}\hfill
    \begin{subfigure}[b]{3cm}
        \centering
        \includegraphics[width=\textwidth]{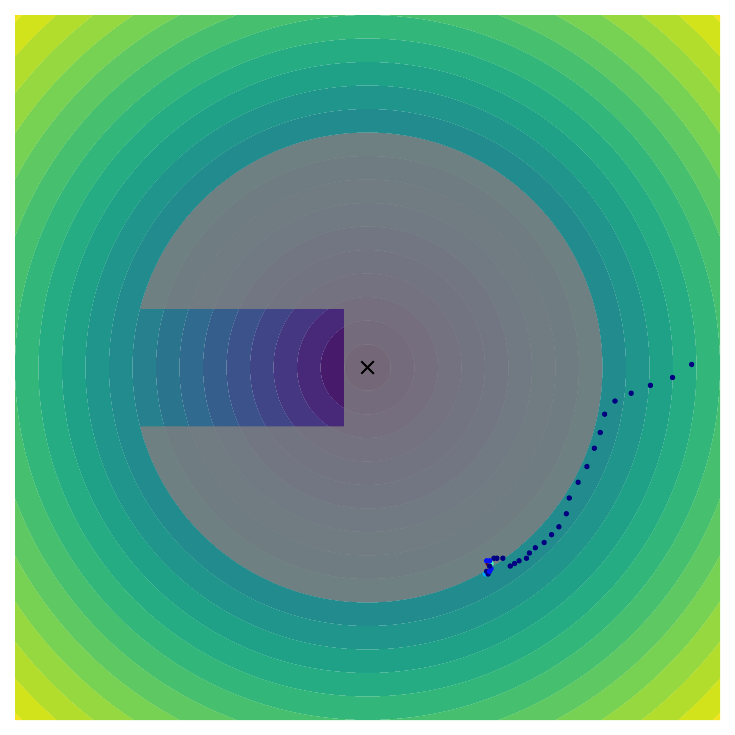}
        \label{fig:SACLBfirstThird}
    \end{subfigure}\hfill
    \begin{subfigure}[b]{3cm}
        \centering
        \includegraphics[width=\textwidth]{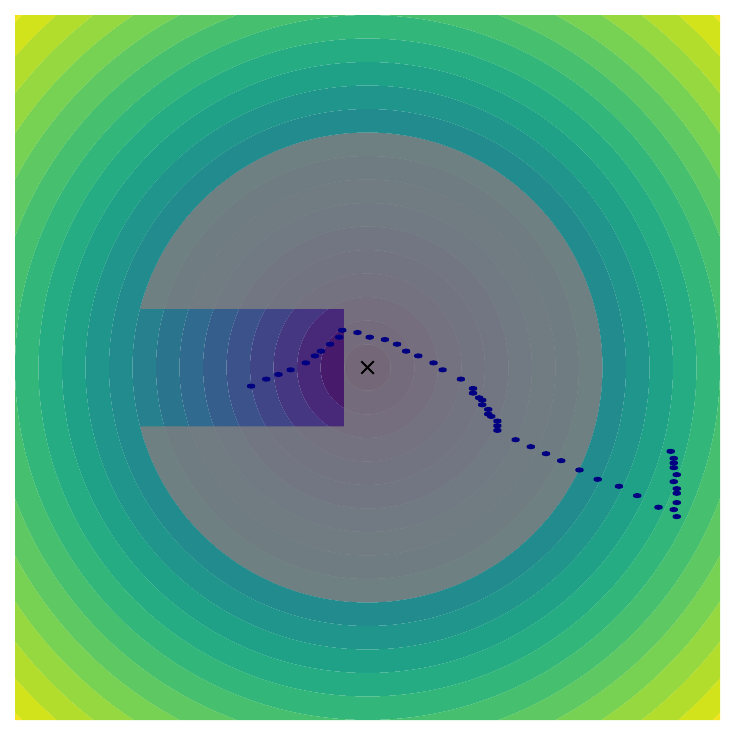}
        \label{fig:WCSACfirstThird}
    \end{subfigure}\hfill
    
    \begin{subfigure}[b]{3cm}
        \centering
        \includegraphics[width=\textwidth]{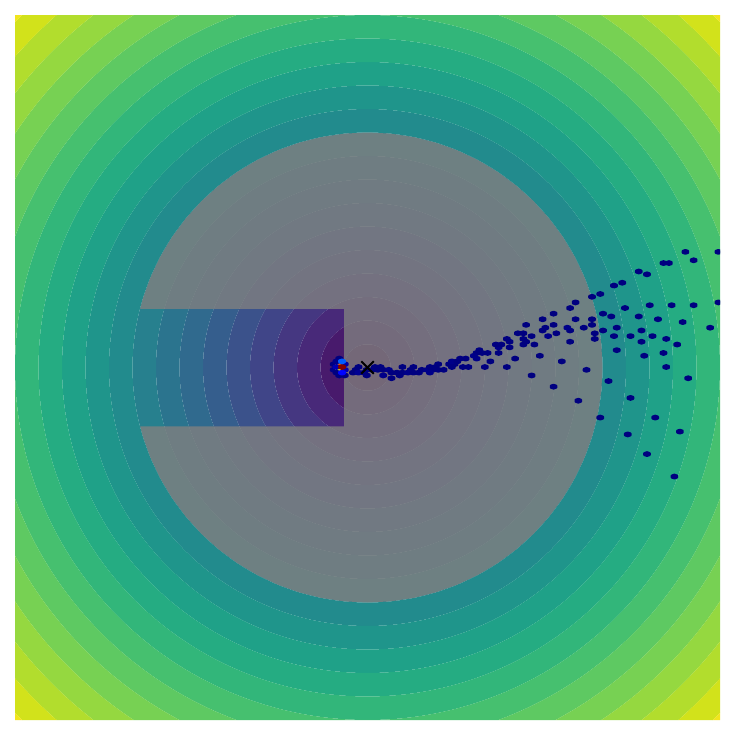}
        \label{fig:TRPOLagsecondThird}
    \end{subfigure}\hfill
    \begin{subfigure}[b]{3.02cm}
        \centering
        \includegraphics[width=\textwidth]{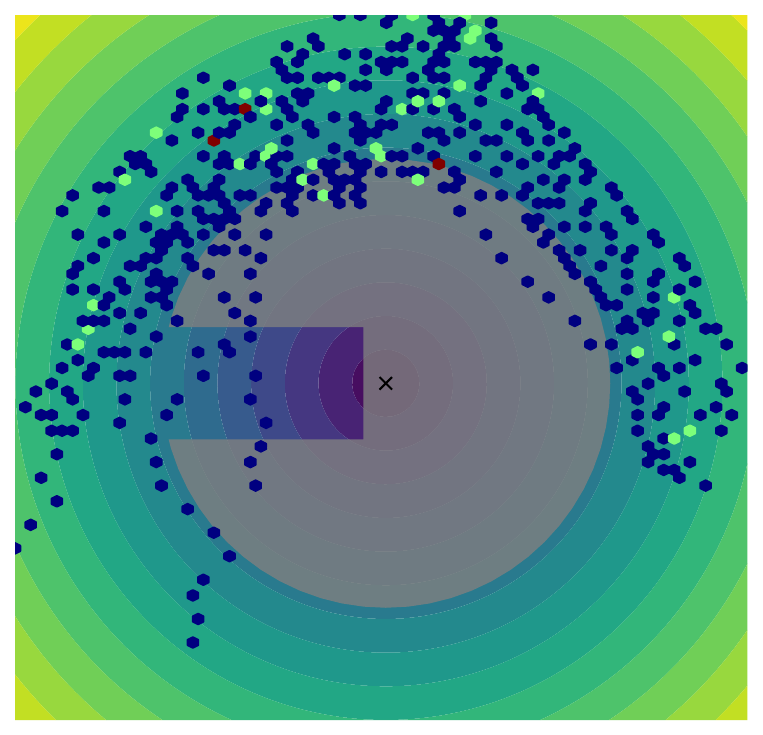}
        \label{fig:CPOsecondThird}
    \end{subfigure}\hfill
    \begin{subfigure}[b]{3cm}
        \centering
        \includegraphics[width=\textwidth]{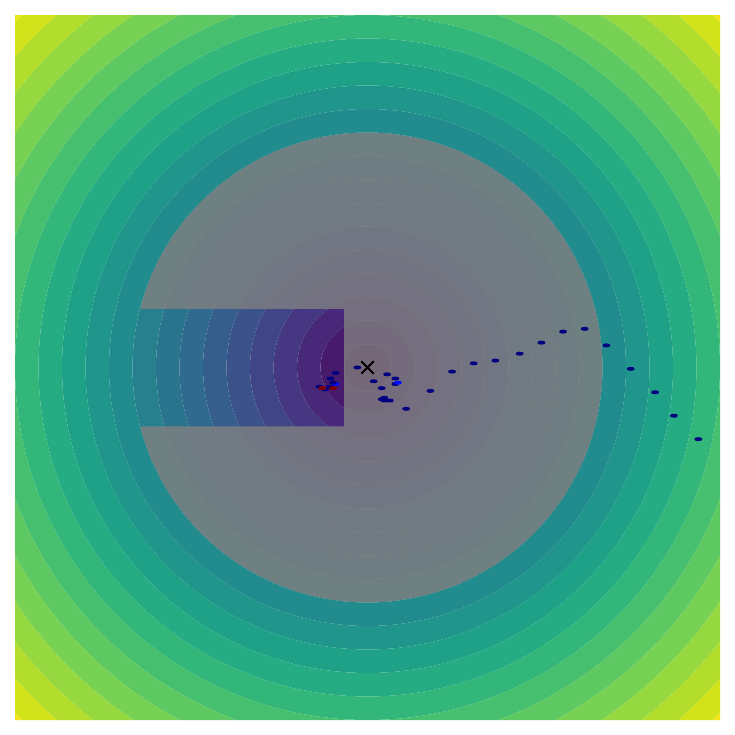}
        \label{fig:SACLagsecondThird}
    \end{subfigure}\hfill
    \begin{subfigure}[b]{3cm}
        \centering
        \includegraphics[width=\textwidth]{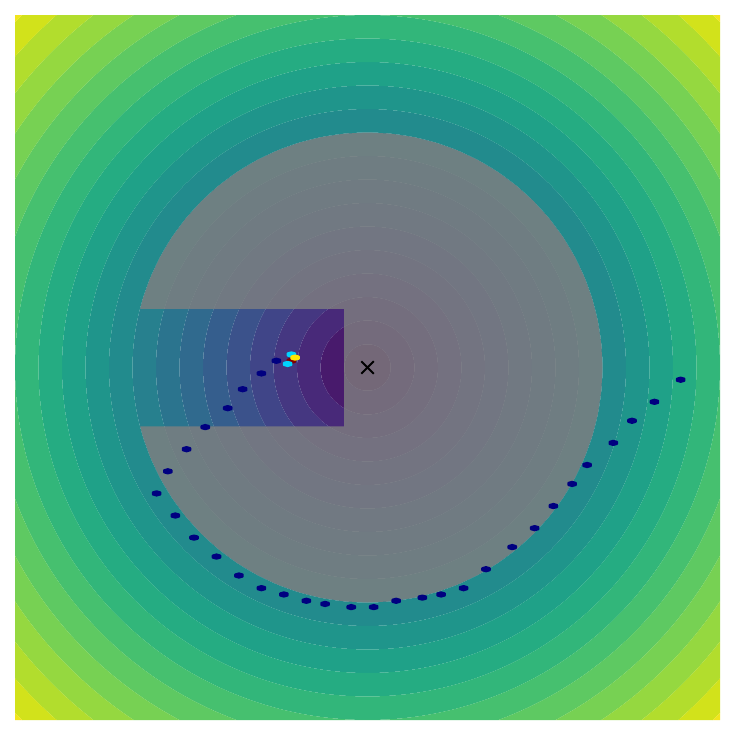}
        \label{fig:SACLBsecondThird}
    \end{subfigure}\hfill
    \begin{subfigure}[b]{3cm}
        \centering
        \includegraphics[width=\textwidth]{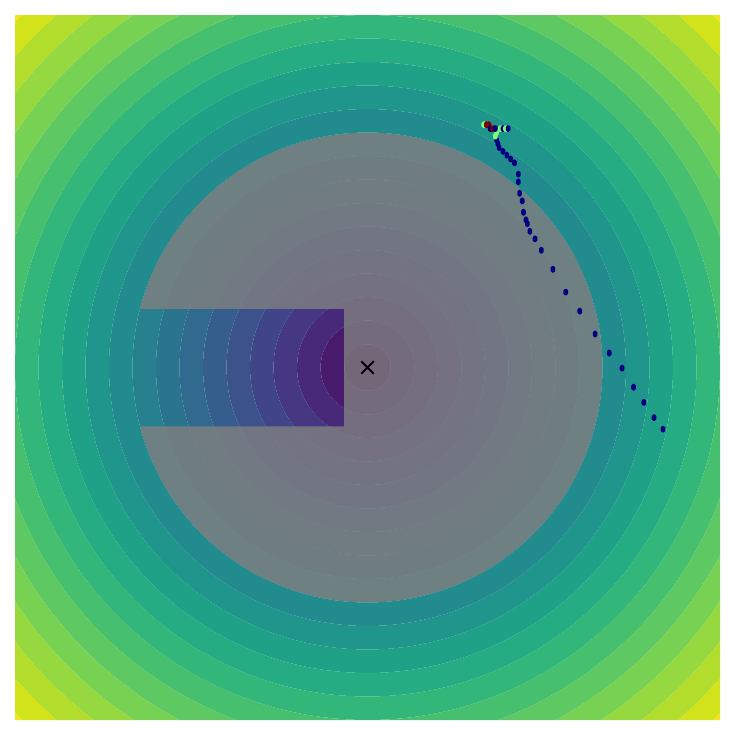}
        \label{fig:WCSACsecondThird}
    \end{subfigure}\hfill
    
    \begin{subfigure}[b]{3cm}
        \centering
        \includegraphics[width=\textwidth]{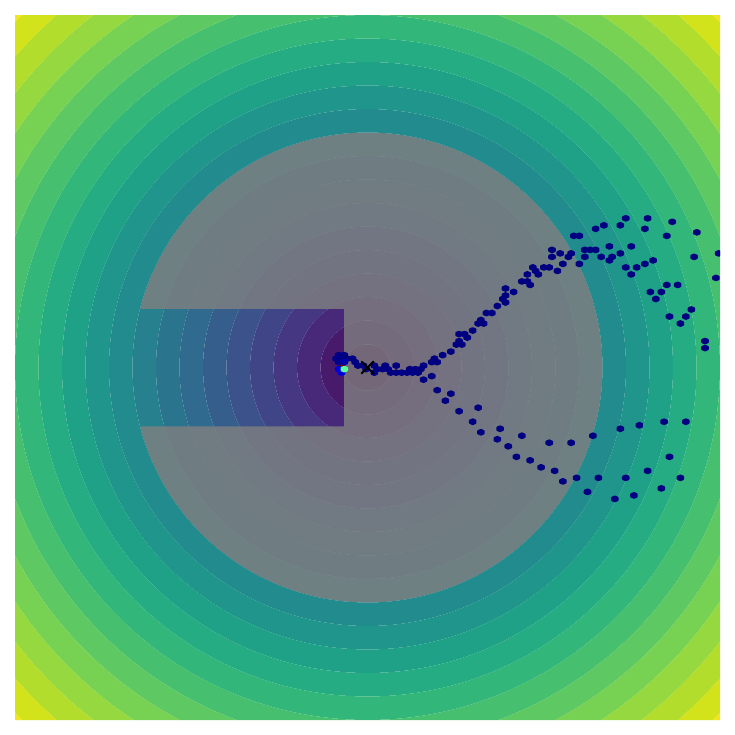}
        \caption{TRPO-Lag}
        \label{fig:TRPOLagthirdThird}
    \end{subfigure}\hfill
    \begin{subfigure}[b]{3.02cm}
        \centering
        \includegraphics[width=\textwidth]{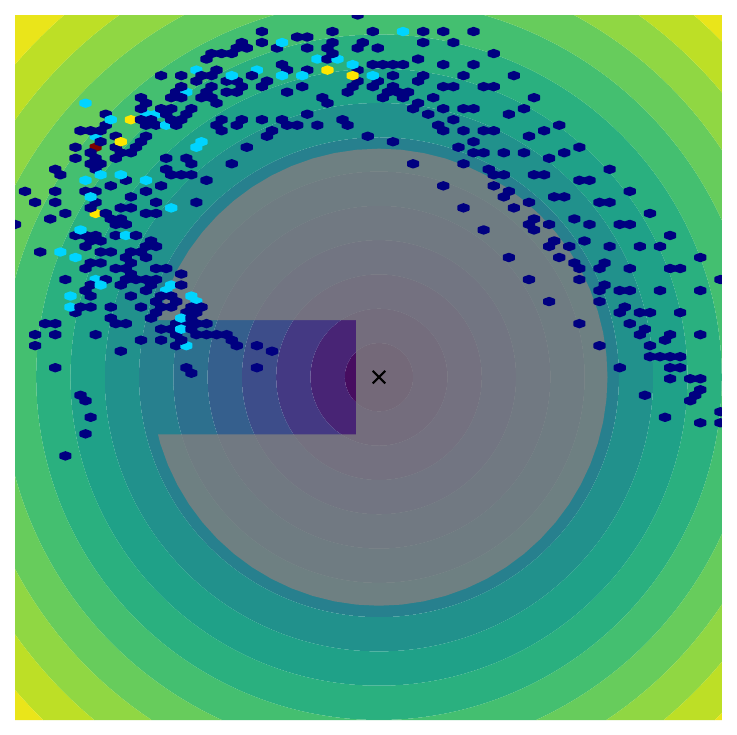}
        \caption{CPO}
        \label{fig:CPOthirdThird}
    \end{subfigure}\hfill
    \begin{subfigure}[b]{3cm}
        \centering
        \includegraphics[width=\textwidth]{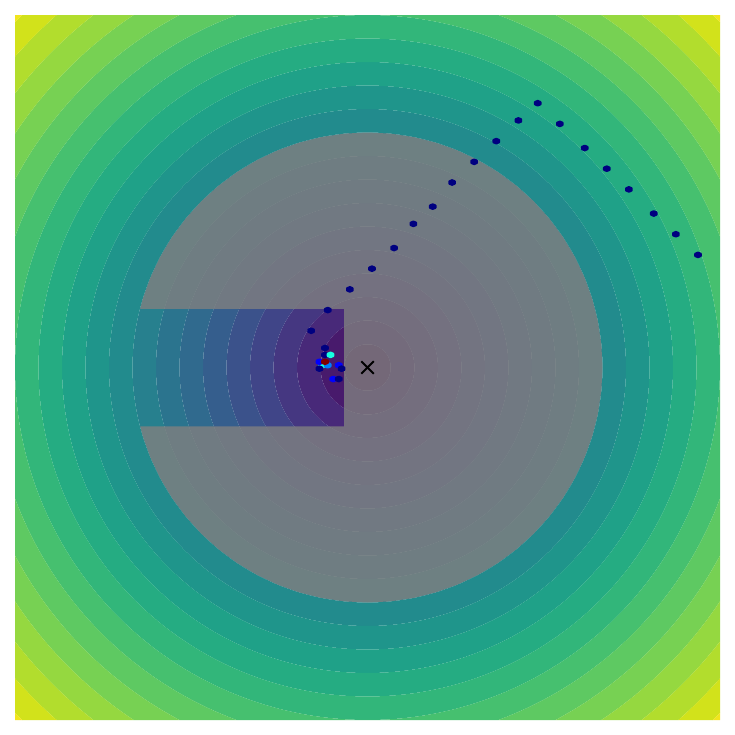}
        \caption{SAC-Lag}
        \label{fig:SACLagthirdThird}
    \end{subfigure}\hfill
    \begin{subfigure}[b]{3cm}
        \centering
        \includegraphics[width=\textwidth]{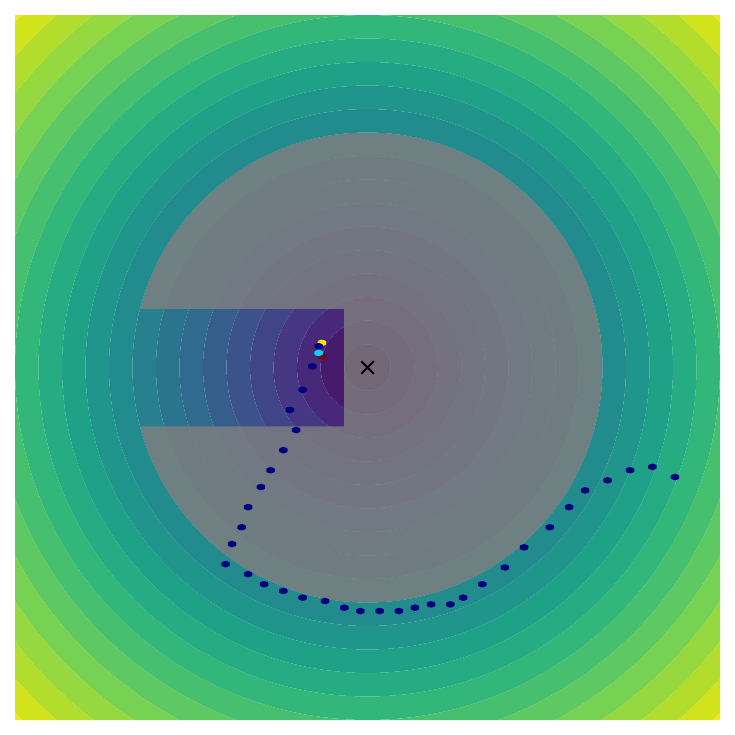}
        \caption{SAC-LB}
        \label{fig:SACLBthirdThird}
    \end{subfigure}\hfill
    \begin{subfigure}[b]{3cm}
        \centering
        \includegraphics[width=\textwidth]{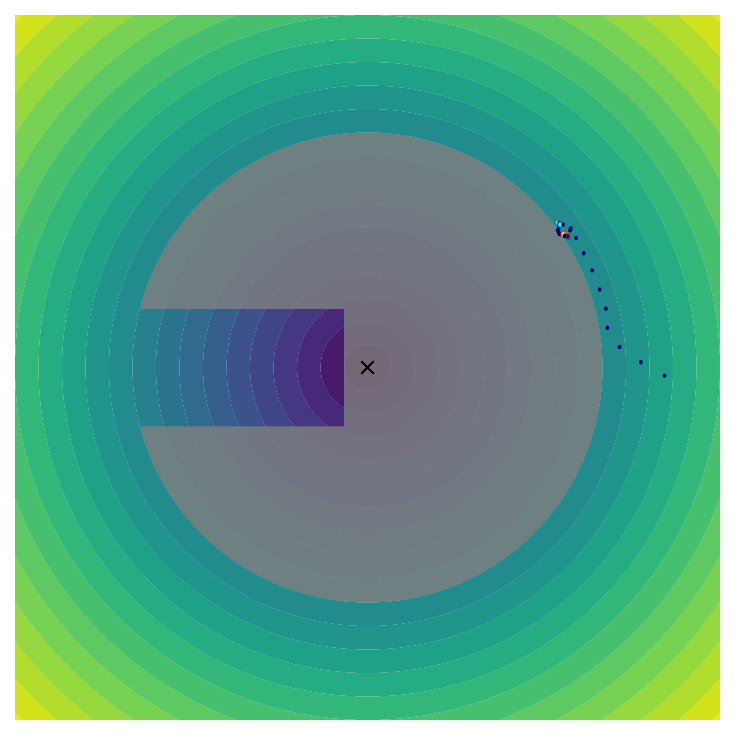}
        \caption{WCSAC}
        \label{fig:WCSACthirdThird}
    \end{subfigure} \\
    \vspace{15pt}
    \caption{Circle2D-1 task heatmaps with rows showing the three training parts. \textbf{First row}: First training stage (0\%-33\% of the training), \textbf{Second row}: Second training stage (33\%-66\% of the training), \textbf{Third row}: Third training stage (66\%-99\% of the training). Heatmaps are chosen as representatives for the exploration behaviour of the algorithms that dominate the EMCC value in their respective training parts. The dots color shows the how frequently a state is visited with dark blue corresponding to single visit and red as most frequently visited. Note that given that off-policy algorithms only collect one trajectory per rollout, the red dots might not be visible easily in the intermediate steps of the rollout. Note that the trajectories always start on the right.}
    \label{fig:Circle2DHeatmaps}
    \vspace{15pt}
\end{figure*}

\paragraph{Results}
The results averaged over 3 seeds for Circle2D tasks are shown in Tab.~\ref{tab:Circle2DResults} and for the Safety-Gymnasium tasks in Tab.~\ref{tab:SafetyGymnasiumResults}.

For the Circle2D-0 tasks we observe that only SAC-Lag and SAC\_LB have runs that end in the left corridor with global feasible optimum. For Circle2D-1 only SAC-LB converges to the left corridor without violating the cost limit. For Circle2D-2 and Circle2D-3 no algorithms manages to safely converge inside the left corridor. Note that the high return values of SAC-Lag and WCSAC for Circle2D-2 are based on strongly varying results for different seeds. On some runs the cost region is ignored and on the others they converge to the local optima with costs of 0, but no runs result in a policy converging inside the left corridor with violating the cost limit.

TRPO-Lag ignores the cost region in levels 1,2,3 and moves straight through it and while CPO explores the boundary of the cost region it fails to converge towards the left corridor. Except for Circle2D-2, as discussed, WCSAC quickly converges towards the local optima with low costs on all other levels.

\paragraph{Analysis}
In general, by looking into EMCC value of different training stages in Tab.~\ref{tab:Circle2DResults} and Tab~\ref{tab:SafetyGymnasiumResults}, we observe a consistent trend across on-policy algorithms in both the Circle2D and Safety-Gymnasium tasks, where EMCC values decrease over training time. This trend indicates that the most safety-critical exploration typically occurs in the early stages of training. In contrast, off-policy algorithms display no clear trend except for occasionally increasing EMCC values towards the end of training, suggesting these algorithms persistently explore safety-critical regions, thus risking severe unsafe behaviors reflected in high EMCC values.

Regarding safe exploration, cost return at evaluation does not adequately measure safety during training. For instance, in the Circle2D-1 task, SAC-Lag shows a lower cost return than on-policy algorithms, yet consistently higher EMCC values by a significant margin. This pattern is also evident in WCSAC and generally across on- and off-policy algorithm comparisons in Circle2D-3 and the SafetyPointCircle task from Safety-Gymnasium. Notably, SAC-Lag often exhibits the highest cost return at evaluation but has the lowest, or nearly the lowest, EMCC values throughout training.

Comparing EMCC with the cost rate metric further highlights its advantages. In the Circle2D-3 task, CPO and WCSAC display identical cost rates but vastly different EMCC values, particularly in the latter stages of training. This discrepancy suggests that cost rate alone may portray an overly simplistic view of safety. For example, the expected cost return curve in Fig.~\ref{fig:Circle2D-3CostReturn} might imply WCSAC undergoes safer training than CPO since it considers also the CVaR value, except for a minor initial spike. However, our analysis underlines that CPO consistently engages in safer exploration compared to WCSAC.

A similar observation applies to the SafetyPointCircle task with TRPO-Lag and SAC-Lag, where despite SAC-Lag's higher cost rate and final evaluation cost return, it shows substantially lower EMCC values. The expected cost return training curve in Fig.~\ref{fig:SafetyPointCircleCostReturn} does not suggest that TRPO-Lag's exploration process is more dangerous than that of SAC-Lag, particularly in the middle and final thirds of the training.

Our results show that TRPO-Lag consistently achieves the lowest EMCC values during the latter stages of training across Circle2D levels 1, 2, and 3. As depicted in Fig.~\ref{fig:Circle2DHeatmaps}, TRPO-Lag tends to bypass the cost region and converge directly towards the optimum, achieving low EMCC values compared to CPO, by finding the shortest path to the optimum. This shortest path to the optimum can be seen in the heatmaps only having high frequency colored dots near the cost region boundary in the left corridor and not inside the cost region. Furthermore the trajectories follow very similar paths and not particularly explore.

However, this behavior can be easily identifiable by analyzing EMCC in conjunction with the expected cost-return curve. Consistently low and stable EMCC values coupled with constant expected cost return values suggest that exploration has ceased since the expected cost return value remains above the cost limit. This can be inferred that the algorithm has converged to local minimum with an unsafe policy.

In Fig.~\ref{fig:Circle2DHeatmaps}, we can also find the different learned policy from different algorithms. Lagrangian-based SafeRL methods tend to ignore the constraints when it is hard to satisfy the constraints, which has also been observed in ~\citep{DBLP:conf/icml/StookeAA20}. WCSAC with CVaR helps with alleviating this issue but fails to explore effectively and converges to a safe yet very conservative policy in terms of the rewards. Despite the imperfect performance in the latter stage of the rollout, SAC-LB with the help of smoothed log barrier function explores the boundary and the resulting policy is closest to the optimal policy by walking along the safe boundary.

\begin{figure}[t]
\centering
\includegraphics[width=0.95\columnwidth]{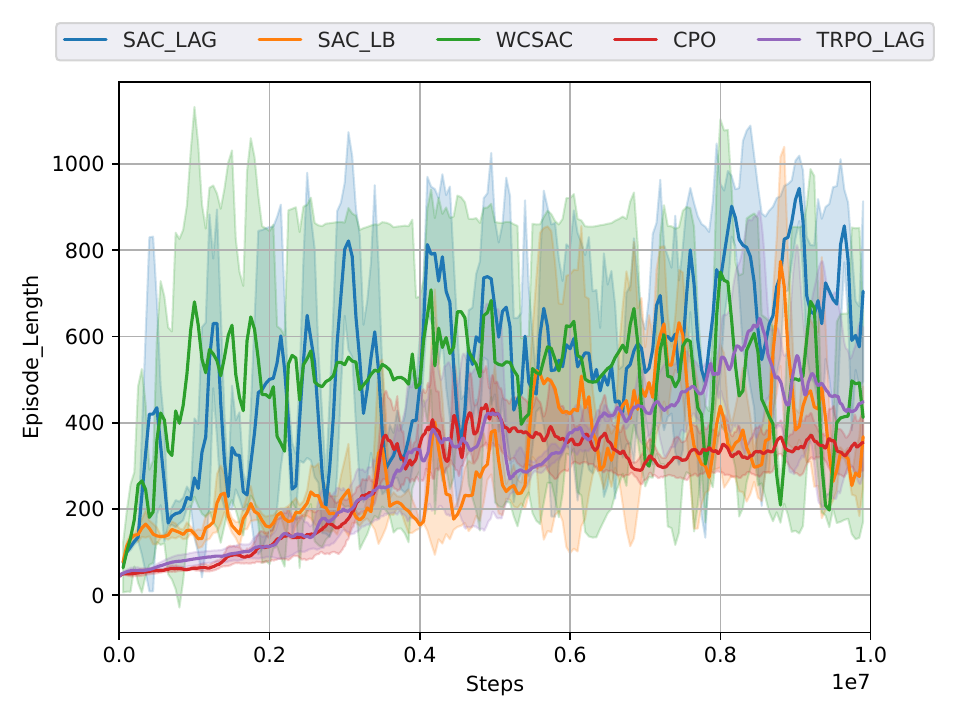}
\caption{Episode length of SafetyHopperVelocity-v1 task over training process. As episodes can terminate early due to the environment the episode length over the training process can vary, exposing additional challenge for benchmarking with conventional metrics as agents might learn to terminate early, resulting in small value of cost of constraint violation.}
\label{fig:SafetyHopperVelocityEpLength}
\vspace{20pt}
\end{figure}

The SafetyHopperVelocity task is the only task that allows for early termination, resulting in variable trajectory lengths during training. Early termination can result in deceptively low values on the expected cost return training curve, potentially misleading observers about the actual safety of the exploration process. Since the Maximum Consecutive Cost (MCC) values are normalized by the respective trajectory length before EMCC calculation, we contend that EMCC can accurately reflect the severity of costs in episodes that terminate early. We prove this claim by comparing the first third of the training process between TRPO-Lag and SAC-LB in the SafetyHopperVelocity task. TRPO-Lag consistently shows lower expected cost return values than SAC-LB (see Fig.~\ref{fig:SafetyHopperVelocityCostReturn}), yet it also features significantly shorter average episode lengths as in Fig.~\ref{fig:SafetyHopperVelocityEpLength}. While the expected cost return curve alone might suggest that TRPO-Lag is safer, incorporating the expected trajectory length reveals a more comprehensive comparison of safe exploration between the two algorithms. EMCC yields a clear outcome, with TRPO-Lag recording a substantially higher EMCC value than SAC-LB, indicating that SAC-LB's exploration is safer in the initial phase of training compared to TRPO-Lag.

\section{Conclusions}
Current metrics used in SafeRL benchmarks either ignore or only allow general impression on safe exploration during the training process. We propose \textbf{E}xpected \textbf{M}aximum \textbf{C}onsecutive \textbf{C}ost steps (EMCC) as metric for assessing safe exploration during training. EMCC allows insights into the different parts of the training and evaluates severity of unsafe actions based on their consecutive occurrence.

We also present a new lightweight benchmark task set, Circle2D, that is tailored for fast evaluation of safe exploration, which also offers standardized interface and parallel training. We believe that this would help SafeRL algorithm designer to rapidly test and evaluate their ideas, thus facilitating the research of community.

As future work, we would propose new objective based on the EMCC metric to encourage the agent to explore safely and effectively.

\begingroup
\begin{table}[h]
\caption{Circle2D environment customization parameters}
\label{tab:EnvironmentConfigParameter}
\centering
    \begin{tabularx}{\columnwidth}{c|c|X} 
    \toprule
    Parameter & Default & Description \\
    \hline
    constraint\_radius & 10.0 & Radius of the circular cost region \\
    init\_radius\_multiplier & 1.5 & Maximum initialization distance given as multiple of constraint\_radius \\
    corridor\_height\_factor & 0.5 & height of the left corridor relative to constraint\_radius. Also resizes local optima cutouts in level 2 and 3. \\
    init\_region\_size & 0.5 & scales the size of the initialization region on the right relative to constraint\_radius\\
    optima\_perturbation & (0,0) & perturbation of the global optimum inside the circular cost region from (0,0) \\
    infeasible\_region\_penetratable & true & flag whether the circular cost region is penetratable \\
    reset\_on\_cost & false & flag whether to reset environment if costs occur \\
    allow\_infeasible\_init & false & whether initilization is allowed inside the cost region \\
    sparse\_reward & false & flag whether only a sparse reward is given when inside a cutout or the left corridor\\
    
    \bottomrule
    \end{tabularx}
\end{table}
\endgroup




\clearpage
\bibliography{mybibfile}

\clearpage
\end{document}